%% file: neurips_2026.tex
\documentclass{article}

\PassOptionsToPackage{numbers, sort&compress}{natbib}

 \usepackage[preprint]{neurips_2026}

\usepackage[utf8]{inputenc} %
\usepackage[T1]{fontenc}    %
\usepackage[colorlinks,linkcolor=red!70!black,citecolor=blue!70!black,urlcolor=blue!70!black]{hyperref}       %
\usepackage{url}            %
\usepackage{booktabs}       %
\usepackage{wrapfig}        %
\usepackage{amsmath}        %
\usepackage{amsfonts}       %
\usepackage{nicefrac}       %
\usepackage{microtype}      %
\usepackage{xcolor}         %
\usepackage{xspace}         %
\usepackage{enumitem}       %
\usepackage{graphicx}
\usepackage{capt-of}

\makeatletter
\renewcommand{\@notice}{\enlargethispage{2\baselineskip}}
\newcommand\blfootnote[1]{%
  \begingroup
  \renewcommand\thefootnote{}\renewcommand\@makefnmark{}%
  \footnotetext{#1}%
  \endgroup
}
\makeatother

\newcommand{\SE}{\ensuremath{\mathrm{SE}(3)}\xspace}
\newcommand{\ours}{MoSE3\xspace}
\newcommand{\pithree}{\ensuremath{\pi^{3}}\xspace}

\title{\ours: Learning World-Space SE(3) at Every Pixel}

\author{%
  Jiahuan Cheng$^{1,3*}$ \:
  Zhiyi Li$^{4*}$ \:
  Tian Xia$^{1*}$ \:
  Ruojin Cai$^{1,2}$ \:
  Yilun Du$^{1,2}$ \:
  Qianqian Wang$^{1,2}$ \\[2pt]
  $^{1}$Harvard University \quad
  $^{2}$Kempner Institute \quad
  $^{3}$Johns Hopkins University \quad
  $^{4}$MIT
}

\begin{document}

\maketitle
\blfootnote{$^{*}$Equal contribution.}

\vspace{-0.5em}
\begin{center}
  \includegraphics[width=\linewidth]{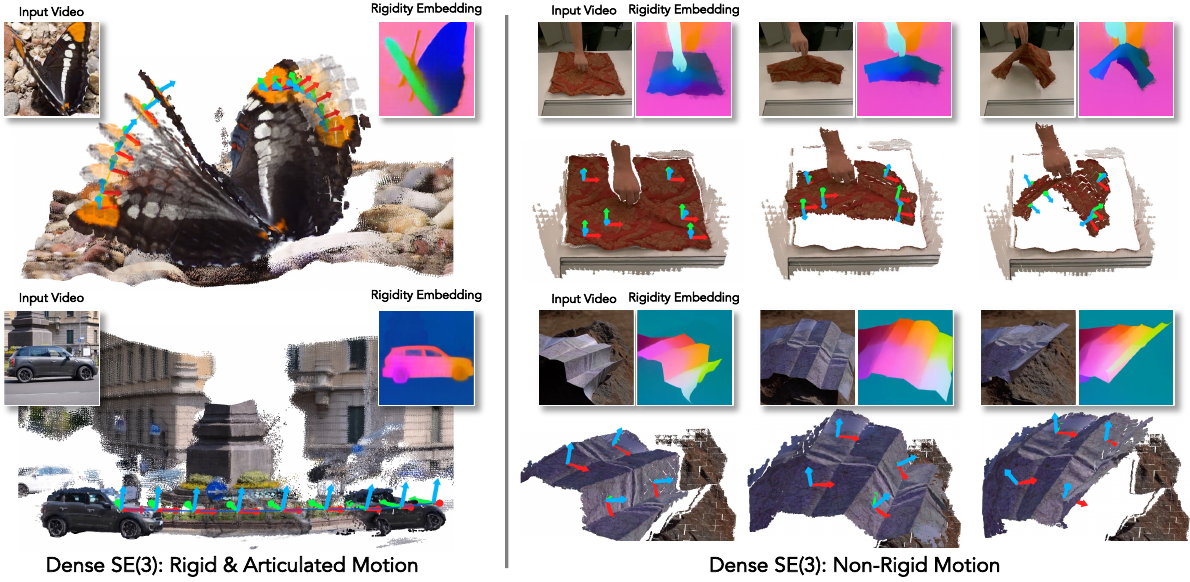}
  \captionof{figure}{\small \textbf{\ours} estimates dense \SE motion from monocular video in a single forward pass. It predicts a 3D rotation and translation at every pixel in a shared world coordinate frame, capturing rigid and articulated motion (left) as well as deformable motion (right). Colored coordinate axes visualize the predicted local transformations at sampled points; insets show the input frames and learned rigidity embeddings.}
  \label{fig:teaser}
\end{center}

\begin{abstract}
  Dense 3D point tracking has been a prominent paradigm for modeling motion in dynamic scenes, but a point track is just a 3-DoF translation curve per pixel: it captures where pixels go, not the rotation of the underlying part, nor which pixels move together as one body. We propose \ours, the first feed-forward model that predicts dense \SE motion from monocular RGB video, producing full 6-DoF rigid transforms at every pixel in world space. Per-pixel \SE motion offers a richer view of how a scene moves: rotation, translation, and grouping all at once. Directly predicting \SE is challenging: rotations lie on a curved manifold that is ill-suited to Euclidean regression, and annotations for \SE are particularly difficult to acquire. To address these challenges, \ours predicts per-pixel \SE through two jointly learned intermediates, 3D point tracks and rigidity embeddings, and recovers \SE by differentiably fitting transforms within each soft rigid cluster, enabling end-to-end prediction and supervision. To close the data gap, we introduce Art-Kubric, a large-scale synthetic dataset with dense \SE and rigidity labels for articulated objects with rich physical interactions. \ours achieves state-of-the-art \SE estimation at pixel, part, and object levels on both rigid and articulated benchmarks, and state-of-the-art average 3D point tracking accuracy across three datasets, while showing strong generalization to real-world videos despite being trained solely on synthetic motion data. \textbf{Project page:} \href{https://mose3-tracker.github.io/}{mose3-tracker.github.io}
\end{abstract}

\input{sections/01_introduction}
\input{sections/02_related_work}
\input{sections/03_method}

\input{sections/04_dataset}
\input{sections/05_experiments}

\input{sections/06_conclusion}

\input{sections/references}
\input{sections/appendix}

\end{document}

%% file: sections/01_introduction.tex
\section{Introduction}
\label{sec:intro}

Understanding how a scene moves over time from video is a foundational problem in computer vision, underpinning applications from physical reasoning to robotic manipulation. The accuracy and richness of the motion understanding bound what downstream systems can do.

Motion estimation from monocular video has advanced rapidly, from short-range optical flow~\cite{horn1981determining, lucas1981iterative, teed2020raft}, to long-range 2D point tracking~\cite{harley2022particle, doersch2022tap, doersch2023tapir, karaev2024cotracker}, to 3D point tracking~\cite{xiao2024spatialtracker, ngo2024delta, zhang2025tapip3d} powered by geometric foundation models~\cite{wang2024dust3r, wang2025vggt, wang2025pi}, each step extending motion estimation across longer horizons and into stronger spatial grounding. Despite this progress, these representations remain low-level: each is a translation curve per pixel, with no rotation and no rigid grouping. However, many downstream tasks need a more structured and compact representation to act on. For example, a robot opening a hinged door wants its rotation axis and joint angle, not the trajectories of a million surface points. The signal exists in current trackers but is buried at the level of individual points.

In this paper, we propose to learn a more structured motion representation from monocular RGB videos: \emph{SE(3) trajectories}, where each pixel carries a full 6-DoF rigid transform in a shared world coordinate frame. A single transform explains the motion of an entire moving part, and the pixels that share it define that part. Deformable objects in turn become smooth fields of locally rigid pieces. This is the representation that physical reasoning, manipulation, and articulated-object understanding need, but directly predicting world-space \SE for in-the-wild RGB video has largely remained out of reach, for several reasons. First, recovering geometry and camera motion for dynamic scenes has itself been an open problem until recently. Second, direct \SE regression generalizes poorly due to the difficulty of its prediction space~\cite{teed2021raft, xiao2024spatialtracker}. Third, long-range scene \SE annotation is exceptionally scarce: even 3D track ground truth is largely confined to synthetic data, and most such datasets lack \SE labels altogether. The prior approaches that recover \SE either operate at the granularity of single rigid objects with known CAD models or depth sensors~\cite{xiang2018posecnn, wen2024foundationpose, wen2023bundlesdf}, require parametric body models~\cite{loper2015smpl, Zuffi:CVPR:2017}, or rely on slow per-video optimization~\cite{wang2024shape, lei2024mosca}.

To address these challenges, we propose \ours, a feed-forward model that predicts per-pixel \SE motion from monocular RGB video. The key insight is to decompose \SE prediction into two intermediates that are individually easier to learn and generalize: dense 3D point tracks, and \emph{rigidity embeddings} that group co-moving pixels into rigid bodies. From these, \SE is recovered analytically by a differentiable weighted Horn~\cite{horn1987closed} fit within each soft rigid cluster. The decomposition lets us leverage the more available 3D tracking supervision  and train the embeddings via the differentiable \SE recovery, making the full prediction end-to-end trainable. To close the data gap, we introduce Art-Kubric, a synthetic dataset and generation pipeline for generating articulated scenes with dense \SE and rigidity labels across a wide range of categories with multi-body physical interactions.

At inference, by clustering the learned rigidity embeddings, \ours recovers object- and part-level 6-DoF poses directly from monocular video. Experiments show \ours achieves state-of-the-art \SE estimation at pixel, part, and object levels on both rigid and articulated benchmarks, and state-of-the-art average 3D point tracking accuracy across three datasets, with strong generalization to real-world videos. 

In summary, our contributions are: \textbf{1)}~\ours, the first feed-forward model for dense \SE motion from monocular RGB video, without category priors, depth sensors, or per-video optimization. \textbf{2)}~A decomposition of \SE into 3D point tracks and rigidity embeddings, connected by a differentiable closed-form fit, enabling end-to-end \SE supervision from scarce annotations and strong generalization. \textbf{3)}~Art-Kubric, a large-scale dataset and generation pipeline with dense \SE and rigidity labels for articulated scenes. \textbf{4)}~State-of-the-art performance for feed-forward \SE estimation at pixel, part, and object levels, and for 3D point tracking, across multiple benchmarks.

%% file: sections/02_related_work.tex
\section{Related Work}

\paragraph{Optical Flow and 2D Point Tracking.}
Optical flow estimates a dense 2D displacement field between consecutive frames~\cite{horn1981determining, lucas1981iterative}. Deep learning~\cite{dosovitskiy2015flownet, ilg2017flownet2, teed2020raft, xu2022gmflow, sun2018pwcnet, wang2024searaft, huang2022flowformer} has driven dramatic accuracy gains, but their predictions are inherently short-horizon.
Long-range point tracking traces back to classical particle-video formulations~\cite{sand2008particle, rubinstein2012towards}. Modern methods~\cite{harley2022particle, doersch2022tap, doersch2023tapir, karaev2024cotracker, karaev2024cotracker3, harley2025alltracker, wang2023tracking, song2024track, lemoing2024dot, doersch2024bootstap, li2024taptr, cho2024local, zholus2025tapnext} have substantially improved long-range 2D tracking accuracy through supervised learning on synthetic data. However, these methods operate in the image plane: their output is a 2D translation trajectory per pixel, with no 3D structure or rigid grouping.

\vspace{0.5em}
\noindent\textbf{Scene Flow and 3D Point Tracking.}
Scene flow extends per-pixel motion from 2D to 3D, with formulations on point clouds~\cite{liu2019flownet3d, gu2019hplflownet} or RGB-D inputs~\cite{teed2021raft, schmid2025ms, wang2024scenetracker}, and more recently on monocular RGB~\cite{yang2020upgrading, liang2025zero}. Long-range 3D point tracking~\cite{zhang2025tapip3d} generalizes this signal across long horizons, predicting 3D trajectories of arbitrary pixels through occlusions. Early methods~\cite{xiao2024spatialtracker, ngo2024delta} lift 2D tracks into 3D using monocular depth~\cite{yang2024depthanything, yang2024depthanythingv2, ke2024marigold, hu2024depthcrafter, bhat2023zoedepth}, but do not disentangle camera from scene motion. More recently, geometric foundation models~\cite{wang2024dust3r,zhang2024monst3r,jiang2025geo4d,leroy2024grounding,wang2025continuous,lin2025depth3} like VGGT~\cite{wang2025vggt} and \pithree~\cite{wang2025pi} estimate cameras and dense geometry in a feed-forward pass. These geometric foundations enable feed-forward world-space 3D tracking, previously only possible through per-video optimization~\cite{wang2024shape, lei2024mosca, yao2025uni4d}. 
Building on pairwise reconstruction models~\cite{wang2024dust3r,leroy2024grounding}, methods like POMATO~\cite{zhang2025pomato}, St4RTrack~\cite{feng2025st4rtrack}, and DPM~\cite{sucar2025dpm} enable world-space 3D tracking. More recent methods~\cite{sucar2025dpm,xiao2025spatialtrackerv2,liu2025trace}, including SpatialTrackerV2~\cite{xiao2025spatialtrackerv2}, Any4D~\cite{karhade2025any4d}, Track4World~\cite{lu2026track4world}, leverage feed-forward multi-view reconstruction models~\cite{wang2025vggt, wang2025pi} to directly predict long-range 3D tracks on continuous videos with improved accuracy and temporal consistency.
However, their primary output remains point translation over time. In contrast, \ours predicts per-pixel \SE motion, capturing rotation, translation, and grouping as a more structured motion representation.

\vspace{0.5em}
\noindent\textbf{\SE Motion Estimation.}
The most common setting for recovering \SE motion is 6-DoF object pose estimation~\cite{wang2019densefusion, brachmann2014learning, xiang2018posecnn, labbe2022megapose, wen2024foundationpose, wen2023bundlesdf,sun2022onepose}, which regresses a single rigid transform per object instance. These methods inherently assume a strictly rigid object, and often require a CAD model~\cite{labbe2022megapose, wen2024foundationpose} or a depth sensor~\cite{wang2019densefusion, wen2024foundationpose}.
For deformable objects, many approaches rely on category-specific priors, from low-rank shape bases~\cite{bregler2000recovering, novotny2019c3dpo} to parametric models like SMPL~\cite{loper2015smpl} and SMAL~\cite{Zuffi:CVPR:2017}. Some 4D object~\cite{yang2021lasr, yang2022banmo} and scene~\cite{lei2024mosca, wang2024shape, lin2025gmc} reconstruction methods can recover local \SE transforms without category priors, but rely on expensive per-video optimization rather than feed-forward prediction. 
On scene-level \SE prediction, piecewise-rigid scene flow~\cite{vogel2013piecewise}, RAFT-3D~\cite{teed2021raft}, and RigidMask~\cite{yang2021rigidmask} estimate per-pixel or per-region rigid motion between frame pairs, while multi-body structure-from-motion~\cite{costeira1998multibody, kumar2016multibody} and dynamic visual odometry~\cite{huang2020clustervo} jointly recover grouping and motion over sparse feature tracks through geometric optimization. Like RAFT-3D, we use learned rigidity embeddings to softly group pixels with shared rigid motion, which yields a finer, part-level decomposition than object-level motion segmentation~\cite{li2015structured, keuper2015motion}. Whereas RAFT-3D iteratively refines pairwise \SE transforms through reprojection-based optimization, we recover world-frame \SE motion through closed-form alignment of predicted 3D point tracks. Earlier learned approaches such as SE3-Nets~\cite{byravan2017se3} and SE3-Pose-Nets~\cite{byravan2018se3posenets} jointly segment scene parts and predict their per-part \SE motion, but operate on point clouds in robotic settings rather than monocular in-the-wild videos. SpatialTracker~\cite{xiao2024spatialtracker} learns rigidity embeddings for feed-forward 3D point tracking, but does not supervise or output per-pixel \SE motion, and its predicted trajectories entangle camera and scene motion.
In summary, \ours is the first to predict per-pixel, world-frame \SE motion feed-forward from monocular RGB video, naturally modeling articulated and deformable objects as collections of local rigidities, without requiring CAD models, depth or stereo input, category templates, or per-video optimization.

%% file: sections/03_method.tex
\begin{figure}[t]
  \centering
\includegraphics[width=\linewidth]
  {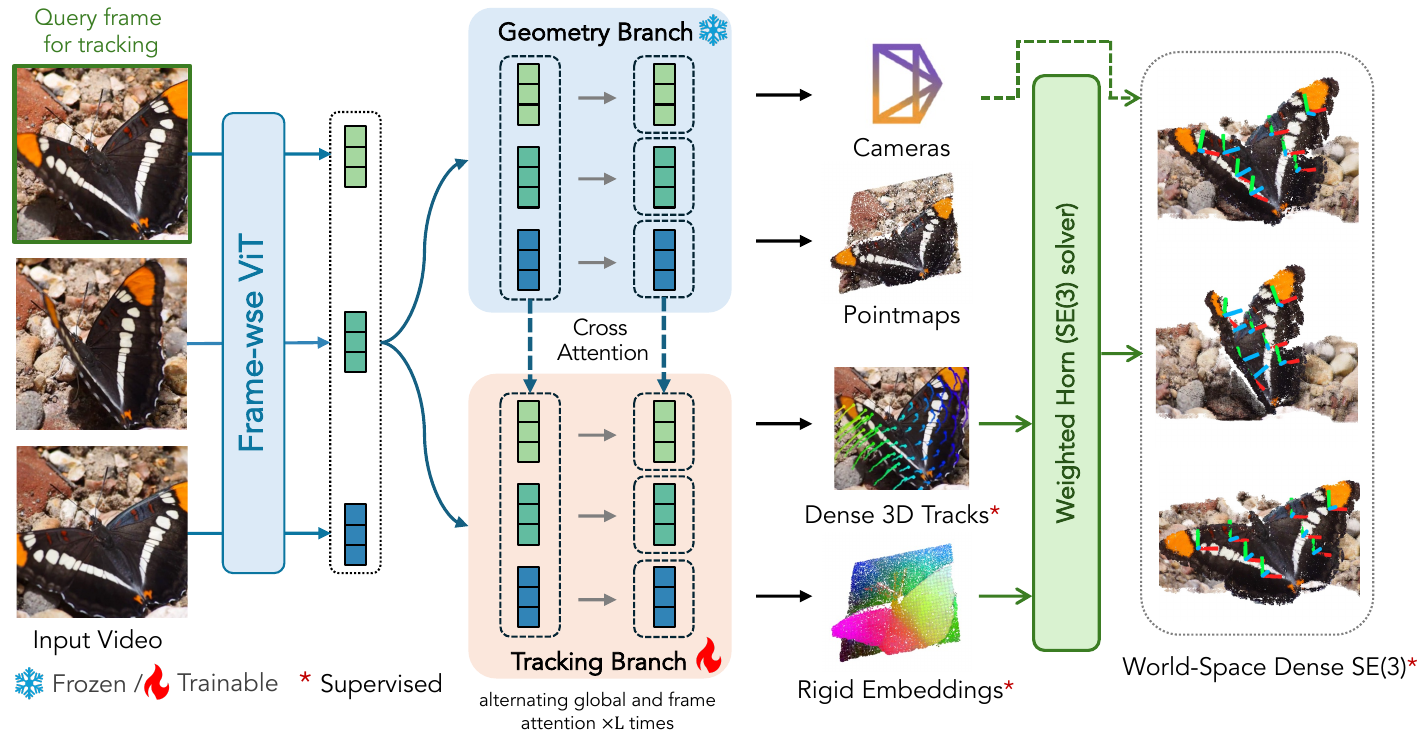}
 \caption{\textbf{Method overview.} \ours extends the frozen \pithree backbone with a trainable tracking branch that predicts dense 3D point tracks $\mathbf{X}^c_i$ and rigidity embeddings $\mathbf{E}_i$. Per-pixel \SE motion is then recovered in closed form via a weighted Horn fit on the world-frame tracks, with weights given by rigidity-embedding similarity. Because the Horn fit is differentiable, the dense \SE prediction is supervised end-to-end together with the tracks and rigidity embeddings.}

  \label{fig:method}
\end{figure}

\section{Method}
\label{sec:method}
We introduce \ours, a feed-forward network that predicts per-pixel \SE motion from a monocular video. Built on the \pithree~\cite{wang2025pi} backbone, \ours introduces two additional predictions: dense point tracks and rigidity embeddings. From these predictions, per-pixel \SE can be recovered in closed form. 
Sec.~\ref{sec:overview} presents our \SE prediction formulation, Sec.~\ref{sec:loss} introduces the training objective, and Sec.~\ref{sec:architecture} describes the model architecture. Fig.~\ref{fig:method} provides an overview.

\subsection{\SE Prediction Formulation}
\label{sec:overview}

Given $N$ input frames $\{I_i\}_{i=1}^N$ at resolution $H \times W$ and a designated query frame index $q \in \{1, \ldots, N\}$ to track from, our goal is to predict a sequence of dense world-frame rigid transformations $\{\mathbf{T}_{q \to i}\}_{i=1}^N$, where each $\mathbf{T}_{q \to i} = (\mathbf{R}_{q \to i}, \mathbf{t}_{q \to i}) \in \SE^{H \times W}$. Let $\mathbf{X}^w_q \in \mathbb{R}^{H \times W \times 3}$ denote the world-frame 3D location of the scene point at each pixel of $I_q$. Then $\mathbf{T}_{q \to i}$ maps $\mathbf{X}^w_q$ to its location at time $i$, $\mathbf{X}^w_i = \mathbf{R}_{q \to i} \mathbf{X}^w_q + \mathbf{t}_{q \to i}$.

A natural approach is to directly regress per-pixel rotation and translation. Although continuous rotation representations support neural regression, learning accurate dense \SE motion remains challenging, particularly when supervision is scarce. We therefore investigate a decomposition that exploits more widely available point-tracking supervision and recovers valid rigid transforms analytically. Our ablation in Sec.~\ref{sec:ablation} compares this formulation with direct \SE regression.

Instead, we make two key observations. First, recent 3D point trackers trained solely on synthetic data generalize well to real-world videos~\cite{feng2025st4rtrack,liu2025trace}. Second, given dense 3D point tracks and a rigid grouping of pixels, \SE motion is recoverable in closed form via Procrustes alignment~\cite{arun1987least,horn1987closed}. Our key insight is thus that predicting \SE can be reduced to predicting two intermediate quantities:  3D point tracks $\mathbf{X}^w_i$, and an L2-normalized \emph{rigidity embedding} $\mathbf{E}_q \in \mathbb{R}^{H \times W \times D}$ trained so that pixels on the same rigid body produce similar embeddings. Both targets are easier to learn and generalize than direct \SE regression (we defer their learning objectives to Sec.~\ref{sec:loss}). From these, the \SE motion of a given pixel $\mathbf{u}$, $\mathbf{T}_{q \to i}(\mathbf{u})$, can be recovered in closed form via Horn's method~\cite{horn1987closed}:
\begin{equation}
\label{eq:se3-opt}
\mathbf{T}_{q \to i}(\mathbf{u}) \;=\; \arg\min_{\hat{\mathbf{T}} \in \SE} \;\sum_{\mathbf{v} \in \mathcal{G}} w(\mathbf{u}, \mathbf{v}) \,\big\| \hat{\mathbf{T}} \cdot \mathbf{X}^w_q(\mathbf{v}) - \mathbf{X}^w_i(\mathbf{v}) \big\|_2^2,
\end{equation}
where the soft weights $w(\mathbf{u}, \mathbf{v})$ are a temperature-$\tau$ softmax over rigidity-embedding inner products,
\begin{equation}
\label{eq:rigid-weights}
w(\mathbf{u}, \mathbf{v}) = \frac{\exp\!\big(\mathbf{E}_q(\mathbf{u})^\top \mathbf{E}_q(\mathbf{v}) / \tau\big)}{\sum_{\mathbf{v}' \in \mathcal{G}} \exp\!\big(\mathbf{E}_q(\mathbf{u})^\top \mathbf{E}_q(\mathbf{v}') / \tau\big)},
\end{equation}
$\mathbf{v}, \mathbf{v}' \in \mathcal{G}$ are sampled reference pixels on the query frame. Intuitively, $w(\mathbf{u}, \mathbf{v})$ measures how rigidly $\mathbf{v}$ moves with $\mathbf{u}$, so the alignment in Eq.~\ref{eq:se3-opt} softly fits $\mathbf{T}_{q \to i}(\mathbf{u})$ to the trajectory of $\mathbf{u}$'s rigid neighbors.
The optimization via weighted Horn is closed-form and differentiable, so \SE supervision flows back into both the point-track and rigidity-embedding predictions.

\subsection{Joint Supervision of Tracks, Rigidity Embeddings, and \SE}
\label{sec:loss}

Given a query frame $q$, our model predicts a 3D point-track map $\mathbf{X}^c_i$ in each target frame $i$'s local camera coordinates, visibility logits $V_i$, and rigidity embeddings $\mathbf{E}_i$. Tracks and visibility are indexed by query pixels, whereas $\mathbf{E}_i$ is defined on frame $i$'s image grid. \pithree additionally predicts pointmaps $\mathbf{M}_i$ and camera-to-world poses $\mathbf{P}_i$, giving $\mathbf{X}^w_i=\mathbf{P}_i\cdot\mathbf{X}^c_i$. We jointly supervise tracking, visibility, rigidity affinities, and the recovered \SE motion.

\subsubsection{Tracking Losses}

We internally parameterize each 3D track position by normalized image coordinates $\mathbf{p}_i=(u_i,v_i)$ and signed depth $z_i$. These are lifted to $\mathbf{X}^c_i$ using focal lengths fitted from \pithree's predicted pointmaps; training uses ground-truth focal lengths for the 3D losses. We supervise normalized image coordinates with an $\ell_1$ loss $\mathcal{L}_{uv}$ and scene-scale-aligned depth with an inverse-depth-weighted $\ell_1$ loss $\mathcal{L}_z$. We additionally apply a query-frame surface-gradient loss $\mathcal{L}_{\nabla}$ and a temporal-displacement loss $\mathcal{L}_{\Delta}$, weighted by $\lambda_{\nabla}$ and $\lambda_{\Delta}$, giving four geometric tracking terms:
\begin{equation}
\label{eq:tracking-loss}
\mathcal{L}_{\mathrm{track}}
=\mathcal{L}_{uv}+\mathcal{L}_z
+\lambda_{\nabla}\mathcal{L}_{\nabla}
+\lambda_{\Delta}\mathcal{L}_{\Delta}.
\end{equation}
The parameterization, lifting, and scale alignment are detailed in Appendix~\ref{sec:appendix_coord_scale}, and the loss definitions in Appendix~\ref{sec:appendix_track_losses}. Separately, a binary cross-entropy loss $\mathcal{L}_{\mathrm{vis}}$ supervises visibility logits $V_i$ against the ground-truth visibility flags on valid annotated tracks.

\subsubsection{Rigidity Embedding Loss}

We supervise rigidity embeddings through two soft-affinity objectives, using \SE annotations when available and 3D track geometry on all eligible clips. Both objectives can apply to the same clip and match a predicted similarity distribution to a row-normalized target affinity. We draw observations across all input frames. An observation $a$ at pixel $\mathbf{u}$ of frame $i$ has embedding $\widetilde{\mathbf{E}}_a=\mathbf{E}_i(\mathbf{u})\in\mathbb{R}^{D}$. Let $\mathcal{J}_a$ hold the other observations $b$ with valid target affinity $A_{ab}$. With temperature $\tau_e$, define
\begin{equation}
\label{eq:affinity-probabilities}
\pi_{ab}=\frac{\exp(\widetilde{\mathbf{E}}_a^\top\widetilde{\mathbf{E}}_b/\tau_e)}
{\sum_{b'\in\mathcal{J}_a}\exp(\widetilde{\mathbf{E}}_a^\top\widetilde{\mathbf{E}}_{b'}/\tau_e)},
\qquad
\pi^\star_{ab}=\frac{A_{ab}}{\sum_{b'\in\mathcal{J}_a}A_{ab'}}.
\end{equation}
Let $\mathcal{O}_A$ contain the observations whose target rows have positive mass. We minimize the mean row-wise divergence
\begin{equation}
\label{eq:affinity-kl}
\mathcal{K}(A)=\frac{1}{|\mathcal{O}_A|}
\sum_{a\in\mathcal{O}_A}\sum_{b\in\mathcal{J}_a}
\pi^\star_{ab}\log\frac{\pi^\star_{ab}}{\pi_{ab}},
\end{equation}
excluding self-pairs and taking $\mathcal{K}(A)=0$ when $\mathcal{O}_A$ is empty. 
Both objectives have this form, differing only in the target affinity $A$. We define its two choices below, and $\mathcal{L}_{\mathrm{emb}}$ combines the resulting losses.

\textbf{\SE affinity loss.}
This term applies only where \SE labels are available. Let $g(a)$ index the motion annotation associated with observation $a$, which can correspond to an object, articulated part, or mesh face. For query-relative ground-truth world transforms $(\mathbf{R}_{i,m}^\star,\mathbf{t}_{i,m}^\star)$, define
\begin{equation}
\label{eq:se3-affinity}
\begin{aligned}
D_{mn}^2&=\max_i\left[
\frac{\operatorname{ang}(\mathbf{R}_{i,m}^{\star\top}\mathbf{R}_{i,n}^\star)^2}{\sigma_R^2}
+\frac{\|\mathbf{t}_{i,m}^\star-\mathbf{t}_{i,n}^\star\|_2^2}{\sigma_t^2}\right],\\
A^{\SE}_{ab}&=\exp\!\left(-D_{g(a)g(b)}^2/\tau_{\mathrm{aff}}\right),
\qquad \mathcal{L}_{\mathrm{aff}}^{\SE}=\mathcal{K}(A^{\SE}).
\end{aligned}
\end{equation}
Here $\sigma_R$ and $\sigma_t$ are angular and translational bandwidths, $\tau_{\mathrm{aff}}$ sets how fast the affinity decays, and $\operatorname{ang}(\cdot)$ is the geodesic angle. Rotations are measured in radians and translations in the original metric world coordinates, before scene-scale normalization. Only annotation trajectories valid across the sampled clip are used. Unlike hard rigid-class targets, these affinities vary continuously with the difference between motion trajectories.

\textbf{Tracking affinity loss.}
This term requires no \SE or rigid-body labels. For ground-truth tracks $a,b$, let $\overline{\mathbf{X}}^{\star}_i$ denote scene-normalized ground-truth positions and $d_i(a,b)=\|\overline{\mathbf{X}}_i^\star(a)-\overline{\mathbf{X}}_i^\star(b)\|_2$. Over their jointly valid frames $\mathcal{V}_{ab}$, define
\begin{equation}
\label{eq:track-affinity}
\Delta_{ab}=\max_{i\in\mathcal{V}_{ab}}d_i(a,b)-\min_{i\in\mathcal{V}_{ab}}d_i(a,b),
\qquad
A^{\mathrm{trk}}_{ab}=\exp(-\Delta_{ab}^2/\sigma_{\mathrm{trk}}^2).
\end{equation}
Here $\Delta_{ab}$ is the spread of the pairwise distance over time, and $\sigma_{\mathrm{trk}}$ its tolerance. Pairs require at least two jointly valid frames. To supervise the per-frame embedding maps, we draw visible track--frame observations. We assign each observation pair the affinity of its underlying tracks and set $\mathcal{L}_{\mathrm{aff}}^{\mathrm{trk}}=\mathcal{K}(A^{\mathrm{trk}})$. This supplies a distance-preservation cue across frames on tracking data with visibility and camera annotations, including clips without \SE labels. Visibility gates the embedding reads, whereas track validity gates the geometric target.

\subsubsection{\SE Losses}
\label{sec:se3-fit}

When \SE ground truth $\mathbf{T}^\star_{q\to i}(\mathbf{u}) = (\mathbf{R}^\star_{q\to i}, \mathbf{t}^\star_{q\to i})$ is available (on Art-Kubric, and post-processed Kubric and Syn4D), we additionally supervise the recovered $\mathbf{T}_{q\to i}(\mathbf{u}) = (\mathbf{R}_{q\to i}, \mathbf{t}_{q\to i})$ directly:
\begin{equation}
\mathcal{L}_{\SE} = \sum_{\mathbf{u}} \Big[ \operatorname{ang}\big(\mathbf{R}_{q\to i}\,\mathbf{R}^{\star\,\top}_{q\to i}\big) + \alpha\,\rho\big(\big\|\mathbf{T}_{q\to i}(\mathbf{u})\cdot\mathbf{c}(\mathbf{u}) - \mathbf{T}^\star_{q\to i}(\mathbf{u})\cdot\mathbf{c}(\mathbf{u})\big\|\big) \Big],
\end{equation}
where $\rho(\cdot)$ is a Huber penalty, $\alpha$ balances the two terms, and $\mathbf{c}(\mathbf{u})$ is the weighted query-frame centroid of the anchor's Procrustes support. The second term compares where the predicted and ground-truth transforms place $\mathbf{c}(\mathbf{u})$.

\subsubsection{Full Loss}

The full objective combines the tracking-branch, rigidity-embedding, and \SE losses described above. On clips that lack \SE or rigidity-embedding labels, the corresponding terms are zeroed. The full multi-task objective is the weighted sum, 
\begin{equation}
\mathcal{L} = \lambda_\text{track}\,\mathcal{L}_\text{track}
            + \lambda_\text{vis}\,\mathcal{L}_\text{vis}
            + \lambda_\text{emb}\,\mathcal{L}_\text{emb}
            + \lambda_{\SE}\,\mathcal{L}_{\SE},
\end{equation}
where all loss weights are fixed. Training runs in a single stage with all losses applied jointly, proceeding in three consecutive phases that differ only in their learning-rate schedule. Full training details, including loss weights and optimization, are in Appendix~\ref{sec:appendix_training}.

\subsection{Geometry-Grounded Tracking Architecture}
\label{sec:architecture}

Having specified what \ours predicts and how it is trained, we now describe the model architecture. We attach a trainable tracking branch alongside \pithree's frozen geometry branch. 
We keep the geometry branch frozen to preserve its pretrained geometric priors while learning motion from tracking data.
The tracking tokens attend to \pithree's geometry tokens at every tracking-branch layer after the shared frozen layers. 
This gives the tracking branch access to a concrete geometric signal throughout the network: predicted tracks in frame $i$ should align with \pithree's pointmap in frame $i$ for co-visible regions.

\noindent\textbf{Tracking Decoder.} Concretely, we reuse \pithree's image encoder and first few decoder layers, then fork into the frozen geometry branch and a trainable tracking branch whose blocks attend with $\mathrm{Q}=\text{tracking tokens}$ and $\mathrm{K},\mathrm{V}=[\text{tracking tokens},\text{geometry tokens}]$. Both task heads below consume the tracking branch's output features.

\noindent\textbf{Point-Tracking Head.} The point-tracking head produces dense $\mathbf{X}^c_i$ and per-pixel visibility logits $V_i$ over all $N$ frames. Its blocks alternate frame-wise and global self-attention and are conditioned on query-frame identity through Adaptive Layer Norm~\cite{peebles2023scalable}, applying one set of normalization parameters to $I_q$'s tokens and another to the rest.

\noindent\textbf{Rigidity-Embedding Head.} The rigidity-embedding head outputs $\mathbf{E}_i$ from the same tracking-branch features. Unlike the point-tracking head, it is query-frame \emph{invariant} (no AdaLN for the query/non-query split), since rigid-body identity is a property of the scene rather than the query. Output embeddings are L2-normalized so that the inner products in Eq.~\ref{eq:rigid-weights} measure cosine similarity.

Additional architecture details, including the layer-wise frozen/trainable split in the tracking branch and the point-tracking head design, are provided in Appendix~\ref{sec:appendix_arch}.

%% file: sections/04_dataset.tex
\section{The Art-Kubric Dataset}
\label{sec:dataset}

We introduce \textbf{Articulated Kubric (Art-Kubric)}, a synthetic dataset and generation pipeline built around six capabilities that, to the best of our knowledge, no existing public corpus combines: (1)~\emph{articulated assets} from $46$ PartNet-Mobility~\cite{xiang2020sapien, mo2019partnet} categories, $226$ procedurally authored Articraft~\cite{zhou2026articraft} categories, and $7$ Infinigen-Articulated~\cite{joshi2025infinigen} categories; (2)~\emph{dense long-range 2D tracks and metric world-space 3D tracks}; (3)~\emph{link-level \SE pose ground truth} for every kinematic link at every frame; (4)~\emph{a per-pixel partition of the image by shared rigid motion}; (5)~\emph{multi-body object interaction} under physics simulation rather than isolated-asset playback, with rigid and articulated bodies sharing every scene; and (6)~a \emph{configurable rendering pipeline} that randomizes HDRI lighting, camera intrinsics, and object composition per scene.

The pipeline lets the community re-render unlimited new clips with the same supervision channels. The release contains $5{,}000$ scenes of $13$--$30$ rigid and articulated objects, about two-thirds of them in motion, dropped into a textured environment and observed by a moving camera, exported with depth, surface normals, three-granularity segmentation, and full camera parameters. Across the release this comes to $163$M tracks, each carrying its rigid-group label, its per-frame occlusion flag, and its link's \SE transform at every frame. Resolution and file layout follow Kubric MOVi-F~\cite{greff2022kubric} so that existing tracking loaders ingest the data without modification. Tab.~\ref{tab:dataset_compare} in the appendix contrasts the supervision channels with prior corpora.

\begin{figure}[t]
  \centering
  \includegraphics[width=\linewidth]{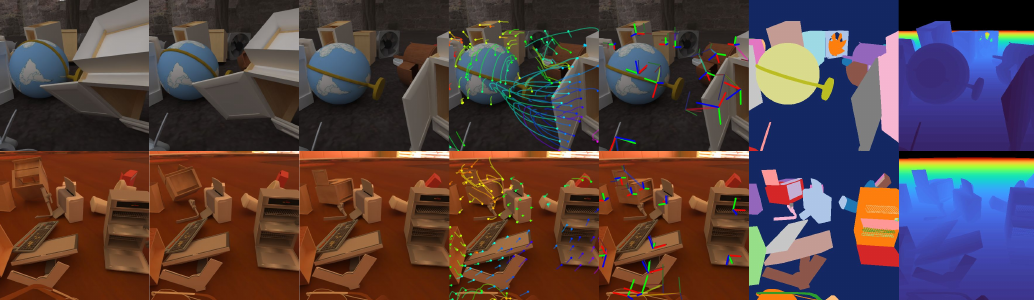}
  \caption{\textbf{Art-Kubric overview.} Two example scenes (one per row). \emph{Columns 1--3:} three RGB frames sampled from the clip. \emph{Columns 4--7 (annotations rendered at the third frame):} sampled long-range 3D tracks colored by their parent rigid body, per-pixel \SE motion as rotated coordinate frames at sparse anchors, rigid-part segmentation (per-pixel partition by shared rigid transform), and depth. Articulated objects from PartNet-Mobility~\cite{xiang2020sapien, mo2019partnet}, Articraft~\cite{zhou2026articraft}, and Infinigen-Articulated~\cite{joshi2025infinigen} interact with rigid Google Scanned Objects~\cite{downs2022google} under physics simulation while a moving camera observes the scene under varying HDRI lighting.}
  \label{fig:dataset_teaser}
\end{figure}

Existing synthetic tracking corpora are either rigid-only, like Kubric~\cite{greff2022kubric, doersch2022tap}, or dominated by skinned humans and animals with no piecewise-rigid parts, like PointOdyssey~\cite{zheng2023pointodyssey} and Dynamic Replica~\cite{karaev2023dynamicstereo}. Rigid-only scenes can provide per-object \SE labels by fitting transforms from tracks and instance masks, but this breaks under articulation: links of one object share a single instance label, so the fitted transform collapses sub-object motion. The rigidity supervision degenerates for the same reason: when every body is a single rigid piece, the partition by shared transform coincides with instance segmentation, and an embedding trained only on such scenes is never asked to split a body into parts. Articulated datasets~\cite{fan2023arctic, liu2022hoi4d, zhao2026synthverse} either cover few object templates or stage one articulated object at a time and drive its joints by sampling angles within their limits, leaving no second body that the embedding has to keep apart. Art-Kubric is the first to combine articulated multi-link motion, physics-driven inter-body contact, and exact per-link \SE ground truth at scale, including for links that are occluded or too small in the image to fit a transform to. Its partition departs from instance segmentation in both directions in every scene: the links of one cabinet fall into separate groups, while every body at rest merges with the floor into one. It provides the articulated \SE supervision that our framework requires, and the ablation in Sec.~\ref{sec:ablation} confirms its contribution to model performance. Full pipeline parameters and the on-disk schema are provided in Appendix~\ref{app:dataset}. We will publicly release the Art-Kubric dataset and generation pipeline.

%% file: sections/05_experiments.tex
\section{Experiments}
\label{sec:experiments}

We evaluate \ours on \SE estimation (Sec.~\ref{sec:eval_se3}) and 3D point tracking (Sec.~\ref{sec:eval_tracking}), and ablate key design choices in Sec.~\ref{sec:ablation}.

\subsection{\SE Evaluation}
\label{sec:eval_se3}
\input{tables/main_se3}

We evaluate on HO3D~\cite{hampali2020honnotate, hampali2022keypointtransformer}, which contains real-world hand-object interactions with rigid objects, and iTACO~\cite{peng2025itaco}, which contains synthetic videos of articulated objects with moving parts. We additionally report results on YCBInEOAT~\cite{wen2020se}, which contains real-world videos of a robot arm manipulating rigid objects, in Tab.~\ref{tab:ycb}. We evaluate at both per-pixel and object/part-level \SE granularities.

To the best of our knowledge, no prior method directly predicts dense world-space \SE motions from monocular videos. We therefore adapt existing 3D tracking methods into baselines using a shared post-hoc protocol. Specifically, for each query pixel, we find its 3D $K$-nearest neighbors at the query frame, treat neighboring pixels as belonging to the same rigid group, and recover a rigid transform from the predicted 3D tracks within that neighborhood via Horn~\cite{horn1987closed}. Since performance depends strongly on the neighborhood size, we sweep $K$ separately for each method on each clip and report the best result. In contrast, \ours predicts a rigidity embedding, allowing us to utilize the learned rigid similarities rather than relying on the hard $k$-NN grouping. We additionally compare against ProxyPose~\cite{zhang2026proxypose}, a 6-DoF pose tracker that needs no such adaptation but predicts a pose per query rather than a dense per-pixel field. Because its video-generation backbone makes many queries expensive, we evaluate its single-query variant and report it only at the object/part level.

For \emph{per-pixel \SE}, we report relative rotation error (RRE, $^\circ$) and ADD~\cite{hinterstoisser2012model} (cm). For \emph{object/part-level \SE}, we cluster pixels into rigid bodies with HDBSCAN~\cite{mcinnes2017hdbscan}. We match each predicted cluster to a ground-truth part mask by IoU and re-fit a single \SE per frame for this part. Methods differ in the clustering feature: \ours clusters directly on the rigidity embedding, while baselines do not predict rigid grouping explicitly. For baselines, we instead form a feature by stacking each pixel's recovered \SE across time and cluster based on that feature. We report the matched cluster's mask IoU together with its RRE, ADD, and AUC$_\text{ADD}$ across the video. See Appendix~\ref{sec:appendix_eval_se3} for additional evaluation details.

\begin{figure}[t]
  \centering
  \includegraphics[width=\linewidth, trim=0 0 0 0, clip]{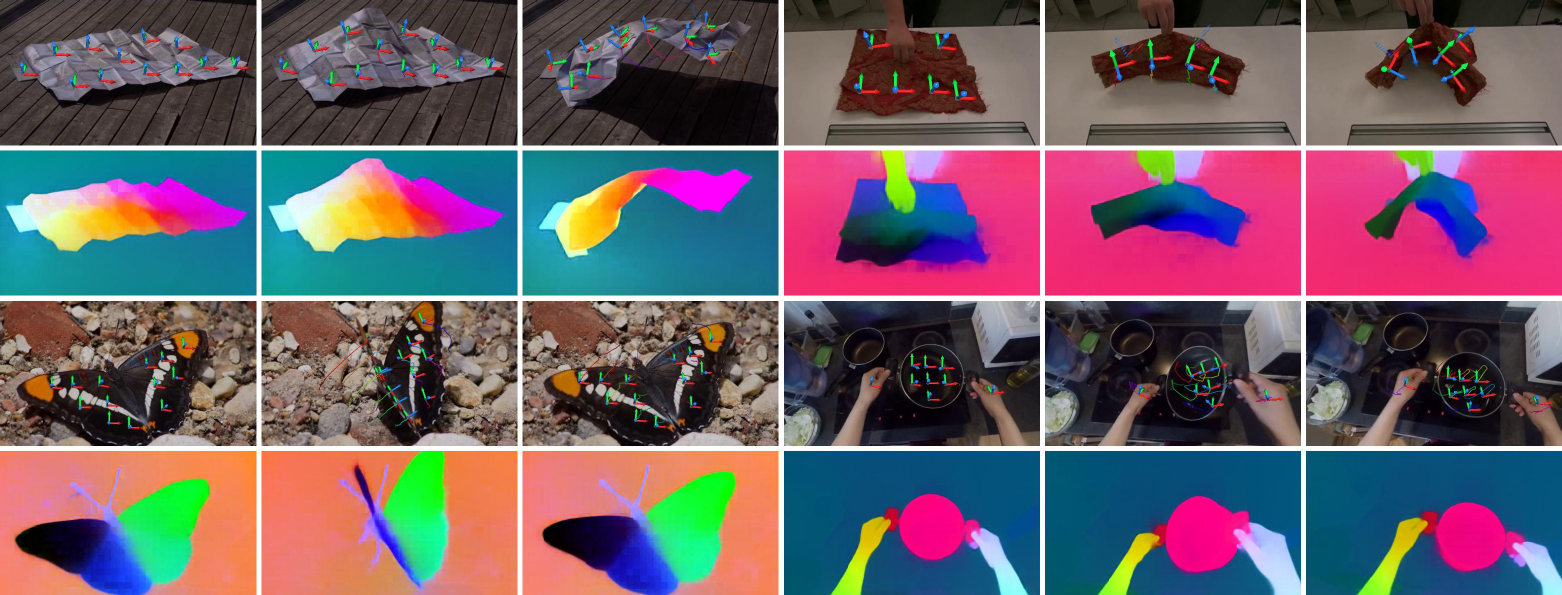}
  \caption{\textbf{Qualitative results on in-the-wild videos.} \ours generalizes to out-of-distribution real-world scenes spanning both rigid and non-rigid cases. For each clip, the three columns are three time steps: the top row shows the predicted per-pixel \SE motion and the bottom row the predicted rigidity embedding.}

  \label{fig:qual}
\end{figure}

Results are shown in Tab.~\ref{tab:main_se3}. \ours achieves the best per-pixel and object/part-level \SE estimation on both rigid (HO3D) and articulated (iTACO) benchmarks across all metrics. At the per-pixel level, \ours consistently reduces rotation error and ADD relative to prior methods, with larger gains on the more challenging iTACO dataset. The improvement carries over to the part and object level, where \ours achieves higher cluster IoU together with lower RRE and ADD. Even when using the same $k$-NN pipeline as the baselines, \ours still outperforms every baseline on every metric, indicating that joint \SE supervision improves the underlying \SE predictions. The rigidity embedding further improves per-pixel \SE and yields substantially higher cluster IoU by grouping rigidly co-moving pixels, providing a more reliable neighborhood for \SE recovery. See Fig.~\ref{fig:qual} for qualitative results on in-the-wild videos.

\subsection{3D Tracking Evaluation}
\label{sec:eval_tracking}
\input{tables/main_tracking}

Following the protocol of Track4World~\cite{lu2026track4world}, we evaluate on PointOdyssey~\cite{zheng2023pointodyssey}, ADT~\cite{pan2023aria}, and PStudio~\cite{joo2015panoptic} from the TAPVid-3D benchmark~\cite{koppula2024tapvid}, reporting tracking accuracy at horizons of 16 and 50 frames (L-16, L-50) in world coordinates. We omit DriveTrack~\cite{balasingam2024drivetrack} as its evaluation split has not been publicly released.

Tab.~\ref{tab:main_tracking} shows that \ours achieves the best average accuracy at both horizons, and the best result on every dataset and horizon except ADT L-16, where it ranks second behind Track4World.
We provide the camera-coordinate tracking comparisons with additional baselines in Appendix~\ref{sec:appendix_additional_results}.

\input{tables/ablation}
\subsection{Ablation Study}
\label{sec:ablation}

We ablate four sets of design choices against the full \ours model (Tab.~\ref{tab:abl_combined}). Due to compute constraints, we run all ablations (including our method) at small scale with identical settings other than the element being ablated. Training setup is provided in Appendix~\ref{sec:appendix_ablation_setup}.

\enlargethispage{\baselineskip}
\noindent\textbf{(a) Architecture.} \emph{Direct \SE regr.} directly regresses \SE, instead of through our track and rigidity-embedding decomposition. This degrades \SE accuracy and generalizes poorly, as direct regression must learn a high-dimensional, manifold-constrained output from scarce \SE labels. Our decomposition reduces the problem to two easier targets with broader supervision coverage, recovering \SE in closed form.

\noindent\textbf{(b) Training data.} Removing Art-Kubric (row \emph{w/o Art-Kubric}) and training only on existing datasets degrades \SE accuracy on iTACO, while on HO3D it leaves \SE accuracy unchanged and lowers only cluster IoU (Tab.~\ref{tab:full_ablation}). Although Kubric already provides dense \SE and rigid-body supervision, its scenes contain only rigid objects with no articulation, which suffices for the single rigid objects of HO3D but not for the articulated, multi-body scenes of iTACO. Art-Kubric closes this gap by adding the multi-object articulated supervision needed to generalize to such scenes.

\noindent\textbf{(c) Training strategy.} Joint training improves \SE accuracy through better tracks, not only better rigid grouping. We train a track-only variant that omits the rigidity-embedding head along with all \SE and rigidity-embedding supervision, then recover \SE post-hoc in two ways: by $k$-NN clustering of its tracks (\emph{Track-only + $k$-NN}), or by grouping its tracks with the rigidity embedding of \emph{Ours (full)} (\emph{Track-only + rigid clust.}). Both underperform \emph{Ours (full)}. The second variant is a controlled comparison: it borrows the full model's embedding, so the grouping weights are identical and only the tracks entering the Procrustes fit differ. The remaining gap is therefore attributable to the tracks alone, showing that supervising tracks jointly with the rigidity and \SE losses makes them more rigid-coherent and in turn yields more accurate \SE.

\noindent\textbf{(d) Loss design.} Starting from the full model, we remove the rigidity loss and the \SE loss one at a time. Both degrade \SE accuracy, with the \SE loss having the larger impact. This confirms that direct end-to-end supervision is the strongest driver of \SE learning. Removing the rigidity loss instead degrades the rigidity embedding more heavily, lowering cluster IoU (Tab.~\ref{tab:full_ablation}): the \SE loss shapes the embedding enough to weight a local Procrustes fit, but not enough to separate rigid bodies globally.

%% file: tables/main_se3.tex
\begin{table}[!t]
\caption{\textbf{\SE estimation on HO3D and iTACO.}
\textit{Per-pixel \SE} evaluates a rigid transform at every pixel; \textit{Object/Part-level \SE} clusters pixels into rigid bodies and recovers one transform per cluster, with IoU measured against ground-truth rigid masks.
ADD is reported in centimeters; AUC$_r$ uses a $30^\circ$ threshold, while AUC$_\text{ADD}$ uses a $10\,\text{cm}$ threshold on HO3D and $20\,\text{cm}$ on iTACO, reflecting the different translation scales of the two benchmarks.
Baselines are evaluated with \textit{$k$-NN}, which clusters predicted point tracks via $k$-NN; for our method we additionally report \textit{rigid clust.}, which clusters our learned rigidity embeddings.
Best in \textbf{bold}, second \underline{underlined}.}
\label{tab:main_se3}
\centering
\scriptsize
\setlength{\tabcolsep}{2pt}
\resizebox{\linewidth}{!}{%
\begin{tabular}{l cccc cccc cccc cccc}
\toprule
& \multicolumn{8}{c}{Per-pixel \SE} & \multicolumn{8}{c}{Object/Part-level \SE} \\
\cmidrule(lr){2-9} \cmidrule(lr){10-17}
& \multicolumn{4}{c}{HO3D} & \multicolumn{4}{c}{iTACO} & \multicolumn{4}{c}{HO3D} & \multicolumn{4}{c}{iTACO} \\
\cmidrule(lr){2-5} \cmidrule(lr){6-9} \cmidrule(lr){10-13} \cmidrule(lr){14-17}
\textbf{Method} & RRE$\downarrow$ & ADD$\downarrow$ & AUC$_r\!\uparrow$ & AUC$_\text{ADD}\!\uparrow$ & RRE$\downarrow$ & ADD$\downarrow$ & AUC$_r\!\uparrow$ & AUC$_\text{ADD}\!\uparrow$ & RRE$\downarrow$ & ADD$\downarrow$ & AUC$_\text{ADD}\!\uparrow$ & IoU$\uparrow$ & RRE$\downarrow$ & ADD$\downarrow$ & AUC$_\text{ADD}\!\uparrow$ & IoU$\uparrow$ \\
\midrule
SpatialTrackerV2~\cite{xiao2025spatialtrackerv2} & 36.45 & 3.73 & 0.376 & 0.633 & 36.71 & 28.45 & 0.408 & 0.309 & 32.56 & 3.46 & 0.670 & 0.385 & 27.25 & 23.01 & 0.421 & 0.239 \\
POMATO~\cite{zhang2025pomato}              & 60.53 & 6.59 & 0.216 & 0.405 & 46.42 & 46.66 & 0.319 & 0.121 & 59.99 & 6.18 & 0.446 & 0.197 & 41.53 & 44.48 & 0.152 & 0.190 \\
ZeroMSF~\cite{liang2025zero}               & 61.53 & 6.70 & 0.224 & 0.409 & 46.52 & 43.68 & 0.298 & 0.128 & 62.29 & 6.43 & 0.442 & 0.230 & 43.47 & 41.83 & 0.158 & 0.187 \\
Any4D~\cite{karhade2025any4d}              & 57.02 & 5.24 & 0.263 & 0.504 & 43.98 & 31.50 & 0.364 & 0.292 & 55.44 & 5.16 & 0.523 & 0.298 & 37.23 & 28.56 & 0.399 & 0.185 \\
V-DPM~\cite{sucar2026vdpm}                 & 40.37 & 3.65 & 0.342 & 0.641 & 32.70 & 22.60 & 0.460 & 0.378 & 38.34 & 3.56 & 0.660 & 0.421 & 28.25 & 19.92 & 0.448 & 0.258 \\
TraceAnything~\cite{liu2025trace}          & 61.56 & 5.69 & 0.245 & 0.467 & 44.73 & 35.47 & 0.347 & 0.249 & 61.20 & 5.65 & 0.488 & 0.221 & 38.45 & 32.47 & 0.341 & 0.142 \\
4RC~\cite{luo20264rc}                      & 38.62 & 3.46 & 0.367 & 0.667 & 28.60 & 20.91 & 0.471 & 0.410 & 36.08 & 3.34 & 0.688 & 0.409 & 18.70 & 17.26 & 0.526 & 0.349 \\
Track4World-Pi3x~\cite{lu2026track4world}  & 33.21 & 3.37 & 0.371 & 0.669 & 28.07 & 24.82 & 0.447 & 0.312 & 30.46 & 3.17 & 0.698 & 0.414 & 18.80 & 20.14 & 0.397 & 0.353 \\
Track4World-DA3~\cite{lu2026track4world}   & 34.61 & 3.34 & 0.361 & 0.671 & 32.09 & 25.69 & 0.408 & 0.321 & 30.98 & 3.20 & 0.694 & 0.383 & 23.78 & 22.07 & 0.399 & 0.319 \\
ProxyPose$^\dagger$~\cite{zhang2026proxypose} & -- & -- & -- & -- & -- & -- & -- & -- & 31.22 & -- & 0.490 & -- & 28.51 & -- & 0.300 & -- \\
\midrule
Ours ($k$-NN)                              & \underline{24.09} & \underline{2.61} & \underline{0.509} & \underline{0.744} & \underline{20.78} & \underline{14.47} & \underline{0.570} & \underline{0.501} & \underline{20.04} & \underline{2.26} & \underline{0.783} & \underline{0.504} & \textbf{9.87} & \textbf{10.27} & \textbf{0.631} & \underline{0.505} \\
\textbf{Ours (rigid clust.)}               & \textbf{18.28} & \textbf{2.16} & \textbf{0.569} & \textbf{0.785} & \textbf{12.30} & \textbf{11.37} & \textbf{0.741} & \textbf{0.594} & \textbf{17.96} & \textbf{2.06} & \textbf{0.802} & \textbf{0.709} & \underline{10.76} & \underline{10.49} & \underline{0.625} & \textbf{0.784} \\
\bottomrule
\end{tabular}
}
\vspace{2pt}
\begin{minipage}{\linewidth}\scriptsize $^\dagger$ProxyPose predicts a pose per query rather than a dense per-pixel field; we evaluate its single-query variant under its own protocol with ground-truth intrinsics, so it is reported at the object/part level only and has no predicted mask for IoU. We omit its ADD, whose mean is dominated by a few clips where its tracker diverges (Appendix~\ref{sec:appendix_eval_se3}).\end{minipage}
\end{table}

%% file: tables/main_tracking.tex
\begin{table}[!t]
\caption{World-coordinate 3D point tracking on PointOdyssey, ADT, and PStudio.
We follow the evaluation protocol of Track4World~\cite{lu2026track4world}.
We report tracking accuracy at horizons L-16 ($\uparrow$) and L-50 ($\uparrow$).
Best in \textbf{bold}, second \underline{underlined}.
Avg.\ is the unweighted mean across the three datasets.}
\label{tab:main_tracking}
\centering
\small
\begin{tabular*}{\linewidth}{l @{\hskip 4pt} @{\extracolsep{\fill}} cc cc cc cc}
\toprule
& \multicolumn{2}{c}{PointOdyssey~\cite{zheng2023pointodyssey}} & \multicolumn{2}{c}{ADT~\cite{pan2023aria}} & \multicolumn{2}{c}{PStudio~\cite{joo2015panoptic}} & \multicolumn{2}{c}{Avg.} \\
\cmidrule(lr){2-3} \cmidrule(lr){4-5} \cmidrule(lr){6-7} \cmidrule(lr){8-9}
Method
& L-16 & L-50
& L-16 & L-50
& L-16 & L-50
& L-16 & L-50 \\
\midrule
SpatialTrackerV2~\cite{xiao2025spatialtrackerv2}                       & 0.1925 & 0.1763 & 0.2456 & 0.2163 & 0.3790 & 0.3689 & 0.2724 & 0.2538 \\
POMATO~\cite{zhang2025pomato}                   & 0.4425 & 0.3905 & 0.3611 & 0.3548 & 0.5166 & 0.4713 & 0.4401 & 0.4055 \\
ZeroMSF~\cite{liang2025zero}                 & 0.4053 & 0.3505 & 0.4530 & 0.3563 & 0.4828 & 0.4386 & 0.4471 & 0.3818 \\
Any4D~\cite{karhade2025any4d}                     & 0.4769 & 0.4174 & 0.4460 & 0.3717 & 0.5707 & 0.5066 & 0.4979 & 0.4319 \\
V-DPM~\cite{sucar2026vdpm}                      & 0.4848 & 0.4233 & 0.4783 & 0.3759 & \underline{0.6084} & \underline{0.5795} & 0.5238 & 0.4596 \\
Track4World~\cite{lu2026track4world}         & \underline{0.5345} & \underline{0.5162} & \textbf{0.6250} & \underline{0.5622} & 0.5946 & 0.5422 & \underline{0.5847} & \underline{0.5402} \\
\textbf{Ours}                          & \textbf{0.5635} & \textbf{0.5650} & \underline{0.5789} & \textbf{0.5953} & \textbf{0.6400} & \textbf{0.6235} & \textbf{0.5941} & \textbf{0.5946} \\
\bottomrule
\end{tabular*}
\end{table}

%% file: tables/ablation.tex
\begin{wraptable}{r}{0.52\linewidth}
\vspace{-40pt}
\caption{\textbf{Ablation study} on iTACO (per-pixel \SE). Trained with a reduced schedule, detailed in Appendix~\ref{sec:appendix_ablation_setup}. ADD in cm; AUC$_r$ at $30^\circ$ and AUC$_\text{ADD}$ at $20\,\text{cm}$. Full results in Tab.~\ref{tab:full_ablation}.}
\label{tab:abl_combined}
\centering
\footnotesize
\setlength{\tabcolsep}{2pt}
\begin{tabular}{l cccc}
\toprule
Variant & RRE$\downarrow$ & AUC$_r\uparrow$ & ADD$\downarrow$ & AUC$_\text{ADD}\uparrow$ \\
\midrule
\multicolumn{5}{l}{\textit{(a) Architecture}} \\
Direct \SE regr.             & 24.57 & 0.525 & 19.05 & 0.393 \\
\midrule
\multicolumn{5}{l}{\textit{(b) Training data}} \\
w/o Art-Kubric               & 22.89 & 0.566 & 17.01 & 0.462 \\
\midrule
\multicolumn{5}{l}{\textit{(c) Training strategy}} \\
Track-only + $k$-NN          & 31.11 & 0.453 & 20.96 & 0.382 \\
Track-only + rigid clust.    & 24.30 & 0.550 & 18.01 & 0.460 \\
\midrule
\multicolumn{5}{l}{\textit{(d) Loss design}} \\
w/o rigidity loss            & \underline{18.97} & \underline{0.614} & \underline{15.83} & \underline{0.480} \\
w/o \SE loss                 & 23.90 & 0.555 & 17.82 & 0.465 \\
\midrule
\textbf{Ours (full)}         & \textbf{17.27} & \textbf{0.647} & \textbf{14.09} & \textbf{0.517} \\
\bottomrule
\end{tabular}
\vspace{-10pt}
\end{wraptable}

%% file: sections/06_conclusion.tex
\section{Conclusion}
\vspace{-10px}
\label{sec:conclusion}

We presented \ours, a feed-forward model that predicts per-pixel \SE motion from monocular RGB video by decomposing \SE into two easier-to-supervise intermediates, dense 3D point tracks and per-pixel rigidity embeddings, with \SE recovered analytically by a differentiable closed-form Horn fit. We also release \emph{Art-Kubric}, a large-scale synthetic dataset that closes the supervision gap for multi-object articulated scenes. \ours achieves state-of-the-art per-pixel and object-level \SE estimation in rigid and articulated scenes, and also sets a new state of the art for 3D point tracking. By turning monocular RGB video into a structured per-pixel motion representation that exposes rotation and rigid grouping alongside translation, \ours provides a foundation for downstream tasks such as physical reasoning, manipulation, and articulated-object understanding.

\paragraph{Limitations.} \ours depends on \pithree for camera geometry, using its camera poses to compose world-frame motion and its pointmaps to fit camera intrinsics, so errors in either propagate into the recovered \SE. Moreover, because \SE is recovered from point tracks and rigidity embeddings rather than regressed directly, accuracy degrades when these intermediates degrade, most notably under fast motion. Finally, we evaluate deformable scenes only qualitatively and leave a quantitative benchmark to future work.

%% file: sections/references.tex
\bibliographystyle{plainnat}
\bibliography{references}

%% file: sections/appendix.tex
\newpage
\appendix
\section*{Appendix}
\renewcommand{\thesection}{\Alph{section}}
\setcounter{section}{0}

\section{Implementation Details}
\label{sec:appendix_training}

\subsection{Training Setup}
\label{sec:appendix_training_detail}

We train \ours by jointly supervising tracks, visibility, rigidity embeddings, and per-pixel \SE transforms.
Training runs for $200$ epochs in a single stage, proceeding in three consecutive phases that differ only in their learning-rate schedule, each resuming from the previous phase's checkpoint.

\paragraph{Initialization.} We initialize \ours from the pretrained \pithree~\cite{wang2025pi} large checkpoint, keeping the \pithree encoder and geometry-branch decoder frozen throughout training. The trainable tracking branch is initialized from the corresponding frozen \pithree weights, with the within-block trainable/frozen split detailed in Sec.~\ref{sec:appendix_tracking_branch}. The transformer layers of both the point-tracking head and the rigidity-embedding head are deep-copied from \pithree's pointmap head transformer layers. The additional global-wise self-attention layers in the point-tracking head are initialized from the corresponding frame-wise self-attention weights. The linear point-track and visibility outputs are deep-copied from \pithree's pointmap and confidence linear heads. Because the point-track head predicts pixel offsets and a depth channel rather than a pointmap, we re-initialize that depth channel so that it predicts unit depth at every pixel at the start of training. Supervision is scale-normalized, so one unit is the mean scene scale of the reconstruction rather than one meter. Only the 16-dimensional rigidity-embedding linear output is trained from scratch.

\paragraph{Optimization.} We optimize with AdamW at peak learning rate $8{\times}10^{-5}$, weight decay $0.05$, and betas $(0.9, 0.999)$, excluding biases and one-dimensional parameters (e.g., LayerNorm) from weight decay. 
For the learning rate schedule, it first warms up over the first epoch ($800$ steps) to $8\times 10^{-5}$. Then, we use a constant learning rate until the validation performance converges. We then gradually decay the learning rate, and use the cosine schedule to $4\times 10^{-8}$ in the last 40 epochs. In total, the model is trained in 200 epochs.
We use per-GPU batch $1$ with no gradient accumulation, yielding an effective batch of $64$ sequences per optimizer step on $64{\times}$ NVIDIA H100-80GB GPUs ($800$ steps per epoch, $160{,}000$ in total). Gradients are clipped at norm $1.0$ and the total loss value is clipped at $100$ to suppress occasional spikes. Mixed precision is applied selectively: all transformer blocks (encoder, decoder layers, and the head transformer layers) run in bf16, while the linear output heads, the post-processing math, and the entire loss module remain in FP32 for numerical stability.

\paragraph{Objective.} We train with $\lambda_\text{track}=1.0$ on both the pixel and depth components of the track loss, $\lambda_\text{vis}=0.1$, $\lambda_\text{emb}=0.1$, and $\lambda_{\SE}=1.0$. Within the track loss, the surface-gradient and temporal-displacement terms use $\lambda_{\nabla}=1.0$ and $\lambda_{\Delta}=2.0$. The rigidity term combines the two affinity losses as $\mathcal{L}_\text{emb}=\mathcal{L}_{\mathrm{aff}}^{\SE}+\tfrac{1}{2}\mathcal{L}_{\mathrm{aff}}^{\mathrm{trk}}$, giving the \SE affinity and tracking affinity losses effective weights of $0.1$ and $0.05$. The \SE affinity loss uses $\tau_e=0.07$, $\tau_{\mathrm{aff}}=5.0$, $\sigma_R=0.12$, and $\sigma_t=0.06$, with up to $4{,}096$ observations sampled per sequence, and the tracking affinity loss uses $\sigma_{\mathrm{trk}}=0.005$. The \SE loss combines a geodesic-angle loss on rotation with a Huber loss ($\delta=0.1$) on the norm of the residual at the anchor's support centroid $\bar{\mathbf{s}}$. The two terms are balanced by $\alpha=20$. Both are computed from the per-anchor weighted Horn fit of Sec.~\ref{sec:appendix_horn}, which uses $4{,}096$ shared anchor and reference points, weights them with $\tau=0.07$ in Eq.~\ref{eq:rigid-weights}, and requires at least $8$ valid points per fit.

\subsection{Per-Anchor \SE Recovery via Weighted Horn}
\label{sec:appendix_horn}

For the closed-form \SE fit of Eq.~\ref{eq:se3-opt}, we use an \emph{anchor} set $\mathcal{A}$ of query-frame pixels for which we predict per-anchor \SE, and a \emph{reference} set $\mathcal{G}$ of query-frame pixels that supply the Procrustes support points. We set $\mathcal{G} = \mathcal{A}$, so every selected query pixel plays a dual role: it is an anchor whose \SE we predict and a Procrustes support point for every other anchor's fit. Both sets are re-sampled per frame pair $(q, i)$ with $i \neq q$ from the query-frame pixels with ground-truth track supervision from $q$ to $i$. We deterministically sub-sample every $k$-th valid pixel in raster order, yielding $|\mathcal{A}| = |\mathcal{G}| = 4{,}096$ with even spatial coverage.

For an anchor $\mathbf{u}\in\mathcal{A}$ and target frame $i$, the rigidity embeddings produce soft weights $\{w_\mathbf{v}\}_{\mathbf{v} \in \mathcal{G}}$ via Eq.~\ref{eq:rigid-weights}, where we abbreviate $w_\mathbf{v} = w(\mathbf{u}, \mathbf{v})$. The anchor is excluded from its own support set, since its unit self-similarity would otherwise dominate the temperature-$\tau$ softmax. Writing the source/target point pairs as $(\mathbf{s}_\mathbf{v}, \mathbf{t}_\mathbf{v}) = (\mathbf{X}^w_q(\mathbf{v}), \mathbf{X}^w_i(\mathbf{v}))$, we solve Eq.~\ref{eq:se3-opt} in closed form by Horn's method~\cite{horn1987closed}. We first form the weighted centroids and the weighted cross-covariance,
\begin{equation}
\bar{\mathbf{s}} = \sum_{\mathbf{v} \in \mathcal{G}} w_\mathbf{v}\,\mathbf{s}_\mathbf{v}, \qquad
\bar{\mathbf{t}} = \sum_{\mathbf{v} \in \mathcal{G}} w_\mathbf{v}\,\mathbf{t}_\mathbf{v}, \qquad
\Sigma = \sum_{\mathbf{v} \in \mathcal{G}} w_\mathbf{v}\,(\mathbf{s}_\mathbf{v} - \bar{\mathbf{s}})(\mathbf{t}_\mathbf{v} - \bar{\mathbf{t}})^\top.
\end{equation}
We use Horn's quaternion representation rather than the SVD-based solution because it empirically provides more stable gradients in this differentiable training loop. We assemble the Horn matrix $H$ from the entries of $\Sigma$; the optimal rotation $\mathbf{R}_{q\to i}(\mathbf{u})$ is the unit quaternion that solves an eigenproblem on $H$, and the optimal translation follows from centroid alignment,
\begin{equation}
\mathbf{t}_{q\to i}(\mathbf{u}) = \bar{\mathbf{t}} - \mathbf{R}_{q\to i}(\mathbf{u})\,\bar{\mathbf{s}}.
\end{equation}
The fit is differentiable, so the resulting per-anchor $\mathbf{T}_{q\to i}(\mathbf{u})$ propagates gradients into both the rigidity-embedding head (through the weights) and the point-tracking head (through the source and reference positions).

\subsection{Coordinate and Scale Conventions}
\label{sec:appendix_coord_scale}

\paragraph{Tracking parameterization and lifting.}
For a query pixel $\mathbf{u}$, the tracking head outputs normalized image coordinates $\mathbf{p}_i=(p_{i,x},p_{i,y})$ and a raw depth parameter $\eta_i$ for every target frame. Let $d=\tfrac12\sqrt{H^2+W^2}$. This $(u,v,z)$ parameterization uses centered image coordinates and decoded depth, not direct XYZ regression or camera-bearing coordinates:
\begin{equation}
\label{eq:uvz-lift}
\begin{aligned}
\mathbf{p}_i&=\left((u_i^{\mathrm{px}}-c_{x,i})/d,\;(v_i^{\mathrm{px}}-c_{y,i})/d\right),
\qquad z_i=\sinh\!\big(\operatorname{clip}(\eta_i,-7,7)\big),\\
\mathbf{X}^c_i&=z_i\begin{bmatrix}d\,p_{i,x}/f_x&d\,p_{i,y}/f_y&1\end{bmatrix}^{\!\top}.
\end{aligned}
\end{equation}
Depth is signed, so valid tracks behind the camera can remain supervised. During training, ground-truth focal lengths $f_{x,i}^\star,f_{y,i}^\star$ replace $f_x,f_y$ when decoding the tracks used by the 3D losses. No ground-truth intrinsics are required at inference.

\paragraph{Focal fitting.}
One focal-length pair is fitted per video from \pithree's frozen pointmap bearings; there is no separately learned intrinsics head. The fit uses image pixel centers $(j+\tfrac12,k+\tfrac12)$ and principal point $(W/2,H/2)$. For the horizontal bearing $b_x$ and centered pixel coordinate $x$, it minimizes $\sum m(b_x-a_xx)^2$ and sets $f_x=1/a_x$; the vertical fit is analogous. Here $m$ selects finite bearings with predicted confidence above $0.1$. Each focal is clamped to $[0.25d,10d]$. Invalid or non-positive estimates, or fewer than $256$ valid samples, trigger a fallback of $d$ on each affected axis. The fitted calibration is detached. Ground-truth principal points are used when projecting ground-truth tracks to read target-frame embeddings for tracking affinity.

\paragraph{Scene normalization and scale alignment.}
Let $\mathcal{R}$ contain valid reconstruction pixels across the sequence. The predicted reconstruction $\mathbf{M}_i$ is in camera $i$'s local frame. The prediction normalizer and the ground-truth normalizer are
\begin{equation}
\label{eq:scene-normalizers}
\begin{aligned}
\widehat\nu&=\frac{1}{|\mathcal{R}|}\sum_{(i,\mathbf{u})\in\mathcal{R}}
\|\mathbf{M}_i(\mathbf{u})\|_2,\\
\nu^\star&=\frac{1}{|\mathcal{R}|}\sum_{(i,\mathbf{u})\in\mathcal{R}}
\| (\mathbf{P}_1^\star)^{-1}\!\cdot\mathbf{M}^{w\star}_i(\mathbf{u})\|_2.
\end{aligned}
\end{equation}
Thus the ground-truth reconstruction norm uses the fixed first-camera frame, whereas the prediction norm uses per-frame local coordinates. Predicted and ground-truth 3D quantities, including camera translations, are divided by their respective normalizers. A further detached scalar $s$ aligns normalized predicted local reconstruction to normalized ground-truth local reconstruction using the inverse-depth-weighted $\ell_1$ scale solver. It uses $4{,}096$ nearest-resampled valid reconstruction points; the absolute solution is clamped to $[0.1,5]$. Define the aligned training tracks
\begin{equation}
\label{eq:aligned-training-tracks}
\widetilde{\mathbf{X}}^c_i=(s/\widehat\nu)\mathbf{X}^c_i,
\qquad
\widetilde{\mathbf{X}}^{c\star}_i=\mathbf{X}^{c\star}_i/\nu^\star.
\end{equation}
The same scalar $s$ is used for depth, surface-gradient, temporal-displacement, and \SE supervision; it is estimated from reconstruction, not from the tracks. UV coordinates are scale invariant. We adopt the per-sequence scale-alignment solver of MoGE~\cite{wang2025moge} (also used in \pithree~\cite{wang2025pi}).

\subsection{Tracking Loss Details}
\label{sec:appendix_track_losses}

\paragraph{UV and depth losses.}
Let $\mathbf{p}_i^\star$ denote the normalized image-coordinate target. Let $\Omega_{uv}$ contain valid tracks with finite predicted and target image coordinates and $\|\mathbf{p}_i^\star\|_2<20$. Let $\Omega_z$ contain valid tracks with finite predicted and target 3D coordinates and finite depth weights. Then
\begin{equation}
\label{eq:uv-depth-losses}
\begin{aligned}
\mathbf{p}_i^\star&=\operatorname{diag}(f_{x,i}^\star/d,f_{y,i}^\star/d)\,
\mathbf{X}^{c\star}_{i,xy}/z_i^\star,\\
\mathcal{L}_{uv}&=\frac{1}{2|\Omega_{uv}|}
\sum_{(i,\mathbf{u})\in\Omega_{uv}}\|\mathbf{p}_i(\mathbf{u})-\mathbf{p}_i^\star(\mathbf{u})\|_1,\\
\mathcal{L}_{z}&=\frac{1}{|\Omega_z|}
\sum_{(i,\mathbf{u})\in\Omega_z}w_i^z(\mathbf{u})
\left|\frac{s}{\widehat\nu}z_i(\mathbf{u})-\frac{z_i^\star(\mathbf{u})}{\nu^\star}\right|.
\end{aligned}
\end{equation}
The depth weight $w_i^z$ is the capped inverse-depth weight of Eq.~\ref{eq:depth-weight}. The UV term is not inverse-depth weighted, and its two coordinates are averaged. Off-screen and behind-camera targets are retained when valid; the radius threshold removes near-zero-depth projection outliers. There is no additional XYZ position loss in the training objective.

\paragraph{Query-frame surface-gradient loss.}
Only the query-frame track map is a surface pointmap on its own image grid. We compare its finite differences against dense ground-truth reconstruction, not against the frozen predicted pointmap. With residual
$\mathbf{r}_q(\mathbf{u})=\widetilde{\mathbf{X}}^c_q(\mathbf{u})-\mathbf{M}^{c\star}_q(\mathbf{u})/\nu^\star$,
let $\mathcal{E}_q$ contain horizontal and vertical adjacent pixel pairs with valid reconstruction and finite predictions and targets at both endpoints. Writing each edge as $(\mathbf{u},\boldsymbol\delta)$, where $\boldsymbol\delta\in\{(1,0),(0,1)\}$,
\begin{equation}
\label{eq:surface-gradient-loss}
\mathcal{L}_{\nabla}=\frac{1}{3|\mathcal{E}_q|}
\sum_{(\mathbf{u},\boldsymbol\delta)\in\mathcal{E}_q}
 w_q^{\nabla}(\mathbf{u})
\|\mathbf{r}_q(\mathbf{u}+\boldsymbol\delta)-\mathbf{r}_q(\mathbf{u})\|_1.
\end{equation}
The weight is taken at the left/top endpoint. This term requires dense query-frame reconstruction ground truth; sparse tracking labels alone do not suffice.

\paragraph{Temporal-displacement loss.}
We compare consecutive sampled-frame displacements in a fixed query-camera frame. Let
$(\mathbf{R}_{qi}^{\pi},\mathbf{t}_{qi}^{\pi})=\mathbf{P}_q^{-1}\mathbf{P}_i$
and $(\mathbf{R}_{qi}^{\pi\star},\mathbf{t}_{qi}^{\pi\star})=(\mathbf{P}_q^\star)^{-1}\mathbf{P}_i^\star$
denote predicted and ground-truth relative cameras before scale normalization. The transported tracks are
\begin{equation}
\label{eq:temporal-transport}
\mathbf{Y}_i=\frac{s}{\widehat\nu}
(\mathbf{R}_{qi}^{\pi}\mathbf{X}^c_i+\mathbf{t}_{qi}^{\pi}),
\qquad
\mathbf{Y}_i^\star=\frac{1}{\nu^\star}
(\mathbf{R}_{qi}^{\pi\star}\mathbf{X}^{c\star}_i+\mathbf{t}_{qi}^{\pi\star}).
\end{equation}
Camera translations are scaled together with the point coordinates. With $\Omega_{\Delta}$ containing consecutive sampled-frame pairs with valid tracks and finite transported points at both endpoints,
\begin{equation}
\label{eq:temporal-displacement-loss}
\mathcal{L}_{\Delta}=\frac{1}{3|\Omega_{\Delta}|}
\sum_{(i,\mathbf{u})\in\Omega_{\Delta}}w_i^{\Delta}(\mathbf{u})
\big\|(\mathbf{Y}_{i+1}-\mathbf{Y}_i)(\mathbf{u})
-(\mathbf{Y}_{i+1}^\star-\mathbf{Y}_i^\star)(\mathbf{u})\big\|_1.
\end{equation}
The weight uses the earlier frame. This matches ground-truth displacement, not zero motion, and is neither a moving-camera-frame difference nor a velocity divided by the sampling interval.

\paragraph{Depth weights and reductions.}
Depth and temporal supervision use capped inverse-depth weights:
\begin{equation}
\label{eq:depth-weight}
\begin{aligned}
a_i(\mathbf{u})&=\max(|z_i^\star(\mathbf{u})|/\nu^\star,10^{-3}),\qquad
\mu_i=\frac{\sum_{\mathbf{u}}m_i(\mathbf{u})a_i(\mathbf{u})}
{\sum_{\mathbf{u}}m_i(\mathbf{u})+HW\cdot10^{-7}},\\
w_i(\mathbf{u})&=\frac{1}{\max(a_i(\mathbf{u}),0.1\mu_i)+10^{-6}}.
\end{aligned}
\end{equation}
For $w^z$, the mask $m$ uses positive-depth valid tracks when any such track exists in the batch, and otherwise all valid tracks. For $w^{\Delta}$, it uses valid finite transported tracks. For $w^{\nabla}$, the same construction uses positive-clamped normalized dense reconstruction depth at $q$ and valid finite surface pixels, rather than absolute track depth. Surface and temporal residuals average their three coordinates. Each loss is divided by the number of valid observations or edges, not by the sum of depth weights. Empty eligible sets contribute zero. The shared UV/depth/visibility helper returns zero when the batch has no valid tracking or reconstruction points.

\paragraph{Visibility supervision.}
Let $\Omega_{\mathrm{vis}}$ contain valid annotated tracks with finite visibility logits, and let $V_i^\star$ be the binary visibility flag. With $\sigma$ the sigmoid,
\begin{equation}
\label{eq:visibility-loss}
\mathcal{L}_{\mathrm{vis}}=-\frac{1}{|\Omega_{\mathrm{vis}}|}
\sum_{(i,\mathbf{u})\in\Omega_{\mathrm{vis}}}
\left[V_i^\star\log\sigma(V_i)+(1-V_i^\star)\log(1-\sigma(V_i))\right].
\end{equation}
Indices are suppressed inside the brackets. No class-balancing or depth weight is applied, and the term is omitted when visibility annotations are absent.

\subsection{Dataset}
\label{sec:appendix_dataset}

For each training sequence, we randomly sample $16$ to $24$ frames at a stride drawn uniformly from $\{1,2,3,4\}$, and load up to $24$ frames of the sampled sequence per GPU. Resolution is randomized per batch, with aspect ratio in $[0.5, 2.0]$ and pixel count in $[100\text{k}, 255\text{k}]$. The query frame index also varies per batch.

The training mix draws from the Kubric CoTracker3 split~\cite{karaev2024cotracker3, greff2022kubric}, Art-Kubric, Syn4D~\cite{jiang2026syn4d}, SynthVerse~\cite{zhao2026synthverse}, PointOdyssey~\cite{zheng2023pointodyssey}, and Dynamic Replica~\cite{karaev2023dynamicstereo}, sampled per sequence at $26\%$, $26\%$, $20\%$, $16\%$, $8\%$, and $4\%$ respectively. Among these, the Kubric CoTracker3 split, Art-Kubric, and Syn4D provide a per-pixel rigid-body partition and per-rigid-body \SE labels. On Syn4D we derive these labels through post-processing, including for deformable objects and human bodies. The \SE affinity loss and the \SE loss are supervised on these three sources only, while the tracking affinity loss needs no such labels and applies to all six. The track and visibility losses are likewise supervised on all six sources.

\section{Architecture Details}
\label{sec:appendix_arch}

\subsection{Tracking Branch}
\label{sec:appendix_tracking_branch}

\pithree-Large has 36 decoder layers. The encoder and decoder layers 0–9 are shared and frozen. Layers 10–35 are paralleled by a 26-layer trainable tracking branch, with each tracking-branch block initialized from its same-layer \pithree decoder counterpart. Specifically, the fused QKV is duplicated into two copies of \pithree's QKV weights: a trainable tracking-side query/key/value triple that processes the tracking tokens, and a frozen geometry-side key/value pair that reads the same-layer geometry tokens from the frozen geometry branch. The remaining components --- LayerNorms, layer scaling, and MLP --- are deep-copies of the \pithree block, with the geometry-side LayerNorms held frozen and the tracking-side trainable. Each block runs a single attention where the queries come from the tracking tokens alone, while the keys and values concatenate the tracking-side and geometry-side projections along the token axis, so one softmax allocates attention between the two streams without an explicit gate.

\subsection{Point-Tracking Head}
\label{sec:appendix_point_tracking_head}

The point-tracking head consists of 10 transformer blocks alternating frame-wise and global self-attention. Each block is conditioned on query-frame identity through Adaptive Layer Norm~\cite{peebles2023scalable}: AdaLN routes each token through one of two learned $(\gamma, \beta)$ pairs based on whether it comes from the query frame $I_q$ or from one of the other $N-1$ frames. Two parallel linear pointmap-style outputs follow the block stack: one predicts the normalized image coordinates $\mathbf{p}_i(\mathbf{u})$ and raw depth parameter $\eta_i(\mathbf{u})$ at every pixel, which Eq.~\ref{eq:uvz-lift} lifts to the 3D track $\mathbf{X}^c_i(\mathbf{u})$, and the other predicts the per-pixel visibility logit $V_i(\mathbf{u})$. $\mathbf{p}_i$ and $\eta_i$ are absolute target-frame values, not offsets from the query pixel.

\section{Evaluation}
\label{sec:appendix_eval}

\subsection{Ablation Setup and Full Results}
\label{sec:appendix_ablation_setup}
\label{sec:appendix_full_ablation}

Tab.~\ref{tab:full_ablation} reports the full ablation across both datasets and both evaluation granularities. The main paper (Tab.~\ref{tab:abl_combined}) reports only per-pixel \SE on iTACO.

All ablations in Tab.~\ref{tab:full_ablation} share an identical training schedule, so any difference in the reported metrics reflects only the variable we change. Each variant is trained as a single stage with all losses enabled, for $80$ epochs of $800$ iterations on $4{\times}$ NVIDIA H200 GPUs, with per-GPU batch $1$ and no gradient accumulation, giving an effective batch of $4$ sequences per optimizer step. We optimize with AdamW at peak learning rate $2{\times}10^{-5}$, following a cosine schedule that decays to $2{\times}10^{-10}$ over training. Dataset mix and loss weights are otherwise identical to those in Sec.~\ref{sec:appendix_training_detail}. For the \emph{Track-only + rigid clust.} variant in group~(c), the track-only model itself does not produce a rigidity embedding, so we cluster on the rigidity embedding from our full-loss model rather than training a separate embedding network.

\input{tables/full_ablation}

\subsection{\SE Evaluation Detail}
\label{sec:appendix_eval_se3}

\paragraph{Ground-truth processing.} We evaluate on 54 clips from HO3D, 55 clips from iTACO, and 9 clips from YCBInEOAT. On HO3D, ground-truth labels are restricted to the B-channel object segmentation, so hand pixels are excluded from all per-pixel and clustering metrics. On iTACO, each clip has multiple actors; we select a single \emph{moving} actor per clip via the largest end-to-start pose difference (matching the iTACO release's heuristic), restrict per-pixel metrics to that actor, and use the binary $\{\text{moving},\,\text{not-moving}\}$ partition as the cluster-IoU GT mask. On YCBInEOAT, each clip contains a single manipulated object with ground-truth pose; we restrict per-pixel and clustering metrics to that object, excluding robot-arm pixels, and use its segmentation mask as the cluster-IoU GT mask. Pixels excluded from these metrics remain in the input — they serve as Procrustes neighbors for baselines and as clustering negatives for \ours. Before computing any metric, we align predicted tracks to ground truth once per clip by fitting a global scale and translation over the pixels that are visible and have ground-truth \SE labels, then apply the same alignment to all predicted points.

\paragraph{Per-pixel \SE.} We compute per-pixel \SE by solving a weighted Horn-Procrustes problem~\cite{horn1987closed} for each labeled query pixel. What differs across methods is the support set and weighting. Baselines have only point tracks, so for each query pixel we take its $K$ nearest 3D neighbors at frame 0 and weight them uniformly. The optimal $K$ depends on scene composition and object scale, so we sweep $K\in\{4,8,16,32,64,128,256\}$ and select per clip the $K$ that minimizes a $z$-score composite of mean RRE, mean ADD, $-$AUC$_r@30^\circ$, and $-$AUC$_\text{ADD}$ at the dataset-scaled ADD threshold. \ours, in contrast, predicts a rigidity embedding that directly identifies which pixels are moving rigidly together. For each query pixel, we weight \emph{every} other pixel by a softmax over its cosine similarity to the query's rigidity embedding at temperature $\tau{=}0.01$, sharper than the $0.07$ used during training, with no explicit neighborhood truncation, so each Procrustes fit pools tracks that the embedding has already grouped as rigidly co-moving. We aggregate the resulting per-pixel transforms over labeled pixels at $t{>}0$ and report mean RRE (degrees), mean ADD (cm), AUC$_r$ up to $30^\circ$, and AUC$_\text{ADD}$ up to a dataset-scaled threshold ($10\,\text{cm}$ on HO3D and YCBInEOAT, $20\,\text{cm}$ on iTACO). ADD is measured on ground-truth surface samples at frame $0$, transformed by the predicted and the ground-truth pose.

For object-level evaluation, we cluster a per-pixel feature with HDBSCAN (min\_cluster\_size $=30$, EOM cluster selection) and match the resulting clusters to the ground-truth mask by maximum IoU. We then re-fit a single \SE per frame on the matched cluster by pooling \emph{all} the tracks it contains. At each target frame $t$, we solve one weighted Procrustes between the frame-0 positions and frame-$t$ positions of every track assigned to the cluster, yielding a single $(\hat{R}_t, \hat{\mathbf{t}}_t)$ that summarizes the object's motion. We construct the per-pixel feature fed to HDBSCAN differently for each method. For \ours, we use the L2-normalized rigidity embedding directly as the cluster feature, since it is trained to group rigidly co-moving pixels, and weight each track in the cluster Procrustes fit by softmax-cosine similarity to the cluster centroid ($\tau{=}0.1$), so tracks more confidently inside the object contribute more. For baselines, we form the feature by stacking each pixel's recovered \SE at two reference frames ($T/2$ and $T{-}1$) into a 24-D vector, and weight all tracks in the cluster fit uniformly. We report mask IoU and the matched cluster's per-frame RRE, ADD, and AUC$_\text{ADD}$, averaged across clips.

\paragraph{ProxyPose.} ProxyPose~\cite{zhang2026proxypose} predicts a pose per query rather than dense per-pixel motion, so we report it only at the object/part level and evaluate it under its own protocol with ground-truth intrinsics. Because its video generation is computationally expensive, we evaluate its one-query variant. We omit its ADD, because its PnP-based proxy-cube tracker diverges on $5$ of $54$ HO3D clips and $7$ of $55$ iTACO clips, producing translations of hundreds of meters that dominate the mean. AUC$_\text{ADD}$ and RRE remain comparable, since diverged clips simply fall outside the AUC threshold and rotation is unaffected.

\input{tables/full_tracking}

\subsection{Additional Results}
\label{sec:appendix_additional_results}

\input{tables/ycb}

Tab.~\ref{tab:full_tracking} reports 3D point tracking results in both camera and world coordinates. The main paper (Tab.~\ref{tab:main_tracking}) reports only world-coordinate results. For comparability with the results reported by Track4World~\cite{lu2026track4world}, we use its released evaluation code unchanged. Before the standard median scaling of TAPVid-3D~\cite{koppula2024tapvid}, this code fits a per-sequence scale and shift to the predicted tracks using only the points that are occluded in the ground truth.

Tab.~\ref{tab:ycb} reports \SE estimation on YCBInEOAT~\cite{wen2020se}, nine real-world clips of a robot arm manipulating rigid objects, following the protocol of Sec.~\ref{sec:eval_se3}. \ours achieves the best result on every metric at both granularities.

\subsection{Qualitative Visualizations}

\begin{figure}[t]
  \centering
  \includegraphics[width=\linewidth, trim=193 107 193 135, clip]{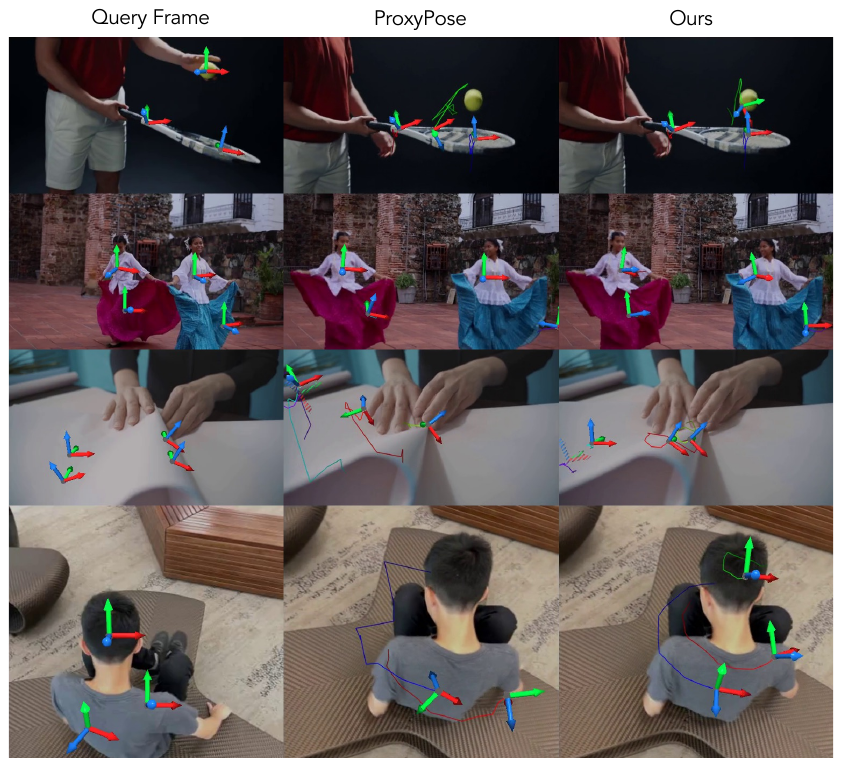}
  \caption{\textbf{Qualitative comparison on in-the-wild videos.} We compare \ours against ProxyPose~\cite{zhang2026proxypose} on out-of-distribution real-world scenes, including deformable cases. For each method, we visualize the estimated per-pixel $\mathbf{T}_{q\to i}(\mathbf{u})$ by drawing its \SE coordinate axes at sampled pixels.}
  \label{fig:comparison}
\end{figure}

\begin{figure}[t]
  \centering
  \includegraphics[width=\linewidth, trim=0 0 0 0, clip]{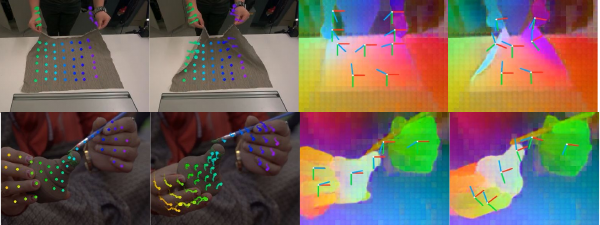}
  \caption{\textbf{Additional qualitative results on in-the-wild videos.} In addition to Fig.~\ref{fig:qual}, we show further out-of-distribution real-world scenes spanning both rigid and non-rigid cases, here also visualizing the predicted 3D point tracks alongside the per-pixel rigidity embeddings and the per-pixel \SE motion overlaid on them.}
  \label{fig:qual_extra}
\end{figure}

Fig.~\ref{fig:comparison} compares \ours against ProxyPose~\cite{zhang2026proxypose} on in-the-wild videos, including deformable cases. \ours produces more stable and coherent axes across both rigid and deformable scenes. We further show additional in-the-wild qualitative results in Fig.~\ref{fig:qual_extra}, complementing Fig.~\ref{fig:qual} in the main paper. The in-the-wild videos in Figs.~\ref{fig:teaser}, \ref{fig:qual}, \ref{fig:comparison}, and \ref{fig:qual_extra} and on our project page are drawn from DAVIS~\cite{pont2017davis}, DyCheck~\cite{gao2022dycheck}, EPIC-KITCHENS-100~\cite{damen2022rescaling}, PhysTwin~\cite{jiang2025phystwin}, AgiBot World~\cite{bu2025agibot}, BridgeData V2~\cite{walke2023bridgedata}, and DROID~\cite{khazatsky2024droid}, as well as from stock footage on Pexels.

\input{tables/dataset}

\section{Art-Kubric: Additional Details}
\label{app:dataset}

\paragraph{Production Setting.} $5{,}000$ scenes at $512{\times}512$, $120$ frames at $60\,$fps, $32{,}768$ tracks per scene with $7\%$ background reserve. Built on the SAPIEN~\cite{xiang2020sapien} simulator with $3{,}018$ articulated URDFs pooled from PartNet-Mobility~\cite{xiang2020sapien, mo2019partnet} ($1{,}034$ models / $46$ categories), Infinigen-Articulated~\cite{joshi2025infinigen} ($1{,}199$ / $7$), and Articraft~\cite{zhou2026articraft} ($785$ / $226$), $1{,}033$ Google Scanned Objects~\cite{downs2022google} as the rigid pool, and HDRI Haven~\cite{greff2022kubric} environment maps ($509$) for image-based lighting plus a tonemapped circular ground projection. Every asset is load-tested with the simulator's own URDF builders before it enters the pool.

\paragraph{Per-Scene Randomization.} Vertical field of view is drawn per scene over $18$--$90^\circ$. Replayed through our loader's crop augmentation, that band reaches below SynthVerse's p05 of $15.9^\circ$ and past PointOdyssey's p95 of $75.4^\circ$, which no single render FOV can, since cropping only narrows a view. Camera radii compensate by $\big(\tan(57.3^\circ/2)/\tan(\mathrm{fov}/2)\big)^{\alpha}$ with $\alpha\sim\mathcal{U}(0.75,1)$ drawn per scene: at $\alpha=1$ the camera distance is an exact function of the field of view ($R^2=0.915$), so a model could read the focal length off apparent object size rather than perspective; the random exponent breaks that dependence ($R^2=0.884$). Distance variety comes instead from the $4$--$7\,$m camera shell. A per-scene lighting level $u\sim\mathcal{U}(0,0.9)$ runs from a strong directional key to a flat environment-only look, and spawn clearance and footprint are drawn below the median object size so that objects can make contact. Each quantity has its own random stream derived from the scene seed, so enabling one leaves the others bit-identical.

\paragraph{Annotations Exported per Scene.} RGB; float32 and uint16 depth; surface normals; three segmentation granularities (\emph{link} / \emph{rigid-part} / \emph{object}); per-link world-frame poses $\mathbf{T}^w_l(t) \in \SE$; oriented 3D bounding boxes; full camera intrinsics and extrinsics; object-coordinate maps (canonical-frame surface positions); 2D and 3D long-range trajectories with occlusion flags. The rigid-part (rp) segmentation collapses links sharing identical rigid motion (with $10^{-3}$ tolerance on translation and rotation) into a single class, so the supervision target for rigid grouping is a partition of pixels by \emph{shared transform}, not by instance.

\paragraph{Statistics.} A typical scene contains $22$ assets ($\sim\!14$ dynamic) decomposed into $\sim\!100$ kinematic links that collapse into $\sim\!85$ rigid-part segments; $90\%$ of scenes contain at least one articulated object, and rigid assets account for $30\%$ of all placed objects. With $32{,}768$ tracks per clip across $5{,}000$ scenes this yields $\sim\!163$M long-range tracks, each carrying a per-pixel rigid-group label and a link-level \SE transform at every one of its $120$ frames.

\paragraph{Extensibility.} The released generation pipeline allows the community to re-render new clips under custom lighting, scenes, and asset distributions; a single dataset-level knob sets the realized share of each asset source, which the pipeline solves for from measured per-source placement rates.

\paragraph{License.} Because some Art-Kubric scenes render PartNet-Mobility objects, whose terms permit only non-commercial research and educational use, we will release the Art-Kubric data (renders, depth, segmentation, tracks, and \SE annotations) under CC BY-NC 4.0. The generation pipeline does not redistribute any upstream asset, so users who re-render scenes obtain each asset source themselves and must comply with its license, namely PartNet-Mobility (non-commercial research and educational use under the ShapeNet terms), Infinigen-Articulated (BSD 3-Clause), Articraft (CC BY 4.0), Google Scanned Objects (CC BY 4.0), and HDRI Haven (CC0).

\section{Broader Impacts}
\label{sec:appendix_broader_impacts}

\ours converts monocular RGB video into a structured per-pixel \SE motion field, with positive applications in 4D reconstruction, robotics, and physical reasoning. Potential negative uses stem from fine-grained motion analysis of humans (e.g., surveillance), but our model outputs geometric motion only and is trained without human identity supervision. We do not foresee additional harms beyond those already present in monocular geometry/tracking models built on \pithree-class foundations.

%% file: tables/full_ablation.tex
\begin{table}[!t]
\caption{\textbf{Full ablation study.}
We ablate four design choices, each compared against our full model (bottom row).
\textbf{(a) Architecture:} directly regressing \SE from an \SE head, instead of predicting rigidity embeddings and point tracks.
\textbf{(b) Training data:} training without Art-Kubric dataset.
\textbf{(c) Training strategy:} predicting tracks alone and recovering \SE posthoc, via either $k$-NN clustering of tracks or clustering of our rigidity embeddings, whose cluster IoU is inherited from our full model and therefore omitted (--).
\textbf{(d) Loss design:} removing individual loss terms.
ADD is reported in centimeters; AUC$_r$ uses a $30^\circ$ threshold, while AUC$_\text{ADD}$ uses a $10\,\text{cm}$ threshold on HO3D and $20\,\text{cm}$ on iTACO.
Best in each column in \textbf{bold}, second \underline{underlined}.}
\label{tab:full_ablation}
\centering
\scriptsize
\setlength{\tabcolsep}{3pt}
\resizebox{\linewidth}{!}{%
\begin{tabular}{l cccc cccc cccc cccc}
\toprule
& \multicolumn{8}{c}{Per-pixel \SE} & \multicolumn{8}{c}{Object/Part-level \SE} \\
\cmidrule(lr){2-9} \cmidrule(lr){10-17}
& \multicolumn{4}{c}{HO3D} & \multicolumn{4}{c}{iTACO} & \multicolumn{4}{c}{HO3D} & \multicolumn{4}{c}{iTACO} \\
\cmidrule(lr){2-5} \cmidrule(lr){6-9} \cmidrule(lr){10-13} \cmidrule(lr){14-17}
Variant & RRE$\downarrow$ & AUC$_r\uparrow$ & ADD$\downarrow$ & AUC$_\text{ADD}\uparrow$ & RRE$\downarrow$ & AUC$_r\uparrow$ & ADD$\downarrow$ & AUC$_\text{ADD}\uparrow$ & RRE$\downarrow$ & ADD$\downarrow$ & AUC$_\text{ADD}\uparrow$ & IoU$\uparrow$ & RRE$\downarrow$ & ADD$\downarrow$ & AUC$_\text{ADD}\uparrow$ & IoU$\uparrow$ \\
\midrule
\multicolumn{17}{l}{\textit{(a) Architecture}} \\
Direct \SE regr.             & 29.79 & 0.449 & 3.65 & 0.655 & 24.57 & 0.525 & 19.05 & 0.393 & 36.18 & 3.21 & 0.701 & 0.600 & 22.49 & 15.01 & 0.511 & 0.620 \\
\midrule
\multicolumn{17}{l}{\textit{(b) Training data}} \\
w/o Art-Kubric               & \textbf{25.07} & \textbf{0.480} & \underline{2.94} & \underline{0.716} & 22.89 & 0.566 & 17.01 & 0.462 & \textbf{23.24} & \underline{2.86} & \underline{0.734} & 0.643 & 19.06 & 15.32 & 0.502 & 0.670 \\
\midrule
\multicolumn{17}{l}{\textit{(c) Training strategy}} \\
Track-only + $k$-NN          & 41.71 & 0.334 & 3.61 & 0.652 & 31.11 & 0.453 & 20.96 & 0.382 & 39.32 & 3.79 & 0.650 & 0.450 & 27.61 & 18.86 & 0.461 & 0.322 \\
Track-only + rigid clust.    & 42.59 & 0.333 & 3.60 & 0.657 & 24.30 & 0.550 & 18.01 & 0.460 & 41.71 & 3.56 & 0.671 & -- & 23.44 & 17.00 & 0.488 & -- \\
\midrule
\multicolumn{17}{l}{\textit{(d) Loss design}} \\
w/o rigidity loss            & 27.38 & 0.456 & \underline{2.94} & \textbf{0.720} & \underline{18.97} & \underline{0.614} & \underline{15.83} & \underline{0.480} & 26.09 & 3.13 & 0.715 & 0.579 & \underline{15.95} & \underline{13.12} & \underline{0.562} & 0.663 \\
w/o \SE loss                 & 43.28 & 0.332 & 3.59 & 0.655 & 23.90 & 0.555 & 17.82 & 0.465 & 41.79 & 3.60 & 0.667 & \underline{0.656} & 21.09 & 15.98 & 0.516 & \textbf{0.702} \\
\midrule
\textbf{Ours (full)}         & \underline{25.53} & \underline{0.473} & \textbf{2.90} & \textbf{0.720} & \textbf{17.27} & \textbf{0.647} & \textbf{14.09} & \textbf{0.517} & \underline{24.57} & \textbf{2.85} & \textbf{0.735} & \textbf{0.689} & \textbf{14.66} & \textbf{12.49} & \textbf{0.564} & \underline{0.685} \\
\bottomrule
\end{tabular}
}
\end{table}

%% file: tables/full_tracking.tex
\begin{table}[!t]
\caption{3D point tracking results on PointOdyssey, ADT, and PStudio in both camera and world coordinates.
We follow the evaluation protocol of Track4World~\cite{lu2026track4world}.
We report tracking accuracy at horizons L-16 ($\uparrow$) and L-50 ($\uparrow$).
Best in \textbf{bold}, second \underline{underlined}.
Avg.\ is the unweighted mean across the three datasets.}
\label{tab:full_tracking}
\centering
\small
\setlength{\tabcolsep}{5pt}
\begin{tabular}{l cc @{\hskip 6pt} cc @{\hskip 6pt} cc @{\hskip 8pt} cc}
\toprule
& \multicolumn{2}{c}{PointOdyssey~\cite{zheng2023pointodyssey}} & \multicolumn{2}{c}{ADT~\cite{pan2023aria}} & \multicolumn{2}{c}{PStudio~\cite{joo2015panoptic}} & \multicolumn{2}{c}{Avg.} \\
\cmidrule(lr){2-3} \cmidrule(lr){4-5} \cmidrule(lr){6-7} \cmidrule(lr){8-9}
Method
& L-16 & L-50
& L-16 & L-50
& L-16 & L-50
& L-16 & L-50 \\
\midrule
\multicolumn{9}{l}{\textit{Camera coordinate 3D tracking}} \\
\midrule
SpatialTracker~\cite{xiao2024spatialtracker}   & 0.3116 & 0.2977 & 0.4962 & 0.4692 & 0.5390 & 0.4991 & 0.4489 & 0.4220 \\
DELTA~\cite{ngo2024delta}                     & 0.3529 & 0.3412 & 0.5116 & 0.4952 & 0.5922 & \underline{0.5533} & 0.4856 & 0.4632 \\
SpatialTrackerV2~\cite{xiao2025spatialtrackerv2}                       & 0.1864 & 0.1785 & 0.2400 & 0.2330 & 0.3784 & 0.3690 & 0.2683 & 0.2602 \\
\midrule
MASt3R~\cite{leroy2024grounding}                   & 0.3546 & 0.3253 & 0.3368 & 0.3029 & 0.3293 & 0.2956 & 0.3402 & 0.3079 \\
MonST3R~\cite{zhang2024monst3r}                 & 0.3912 & 0.3860 & 0.3694 & 0.3429 & 0.3511 & 0.3381 & 0.3706 & 0.3557 \\
POMATO~\cite{zhang2025pomato}                   & 0.4816 & 0.4623 & 0.5338 & 0.5299 & 0.5163 & 0.4726 & 0.5106 & 0.4883 \\
ZeroMSF~\cite{liang2025zero}                 & 0.4214 & 0.3887 & 0.5382 & 0.4635 & 0.5083 & 0.4524 & 0.4893 & 0.4349 \\
Track4World~\cite{lu2026track4world}         & \underline{0.5397} & \underline{0.5268} & \textbf{0.6501} & \underline{0.6091} & \underline{0.5948} & 0.5423 & \underline{0.5949} & \underline{0.5594} \\
\textbf{Ours}                          & \textbf{0.5666} & \textbf{0.5688} & \underline{0.5951} & \textbf{0.6146} & \textbf{0.6391} & \textbf{0.6207} & \textbf{0.6003} & \textbf{0.6014} \\
\midrule
\multicolumn{9}{l}{\textit{World coordinate 3D tracking}} \\
\midrule
SpatialTrackerV2~\cite{xiao2025spatialtrackerv2}                       & 0.1925 & 0.1763 & 0.2456 & 0.2163 & 0.3790 & 0.3689 & 0.2724 & 0.2538 \\
POMATO~\cite{zhang2025pomato}                   & 0.4425 & 0.3905 & 0.3611 & 0.3548 & 0.5166 & 0.4713 & 0.4401 & 0.4055 \\
ZeroMSF~\cite{liang2025zero}                 & 0.4053 & 0.3505 & 0.4530 & 0.3563 & 0.4828 & 0.4386 & 0.4471 & 0.3818 \\
Any4D~\cite{karhade2025any4d}                     & 0.4769 & 0.4174 & 0.4460 & 0.3717 & 0.5707 & 0.5066 & 0.4979 & 0.4319 \\
V-DPM~\cite{sucar2026vdpm}                      & 0.4848 & 0.4233 & 0.4783 & 0.3759 & \underline{0.6084} & \underline{0.5795} & 0.5238 & 0.4596 \\
Track4World~\cite{lu2026track4world}         & \underline{0.5345} & \underline{0.5162} & \textbf{0.6250} & \underline{0.5622} & 0.5946 & 0.5422 & \underline{0.5847} & \underline{0.5402} \\
\textbf{Ours}                          & \textbf{0.5635} & \textbf{0.5650} & \underline{0.5789} & \textbf{0.5953} & \textbf{0.6400} & \textbf{0.6235} & \textbf{0.5941} & \textbf{0.5946} \\
\bottomrule
\end{tabular}
\end{table}

%% file: tables/ycb.tex
\begin{table}[!t]
\caption{\textbf{\SE estimation on YCBInEOAT~\cite{wen2020se}.}
Nine clips of robot manipulation with rigid objects, evaluated with the protocol of Tab.~\ref{tab:main_se3}.
ADD is reported in centimeters; AUC$_r$ uses a $30^\circ$ threshold and AUC$_\text{ADD}$ a $10\,\text{cm}$ threshold.
Best in \textbf{bold}, second \underline{underlined}.}
\label{tab:ycb}
\centering
\scriptsize
\setlength{\tabcolsep}{3pt}
\begin{tabular}{l cccc cccc}
\toprule
& \multicolumn{4}{c}{Per-pixel \SE} & \multicolumn{4}{c}{Object/Part-level \SE} \\
\textbf{Method} & RRE$\downarrow$ & ADD$\downarrow$ & AUC$_r\uparrow$ & AUC$_\text{ADD}\uparrow$ & RRE$\downarrow$ & ADD$\downarrow$ & AUC$_\text{ADD}\uparrow$ & IoU$\uparrow$ \\
\midrule
SpatialTrackerV2~\cite{xiao2025spatialtrackerv2} & 21.93 & 4.93 & 0.461 & 0.576 & 22.93 & 6.55 & 0.508 & 0.251 \\
POMATO~\cite{zhang2025pomato}              & 46.47 & 10.78 & 0.223 & 0.336 & 50.11 & 10.89 & 0.309 & 0.054 \\
ZeroMSF~\cite{liang2025zero}               & 44.79 & 11.57 & 0.243 & 0.353 & 46.71 & 10.74 & 0.387 & 0.073 \\
Any4D~\cite{karhade2025any4d}              & 42.30 & 10.28 & 0.251 & 0.432 & 44.67 & 10.45 & 0.455 & 0.267 \\
V-DPM~\cite{sucar2026vdpm}                 & 30.42 & 7.58 & 0.329 & 0.504 & 31.60 & 8.18 & 0.501 & 0.234 \\
TraceAnything~\cite{liu2025trace}          & 44.79 & 9.61 & 0.249 & 0.415 & 48.47 & 10.18 & 0.392 & 0.066 \\
4RC~\cite{luo20264rc}                      & 21.18 & 4.85 & 0.492 & \underline{0.614} & 24.23 & 4.52 & 0.625 & 0.322 \\
Track4World-Pi3x~\cite{lu2026track4world}  & 19.92 & 7.68 & 0.441 & 0.562 & 19.91 & 7.53 & 0.549 & \underline{0.360} \\
Track4World-DA3~\cite{lu2026track4world}   & 22.46 & 5.59 & 0.399 & 0.560 & 28.89 & 6.11 & 0.542 & 0.278 \\
ProxyPose$^\dagger$~\cite{zhang2026proxypose} & -- & -- & -- & -- & 17.01 & -- & 0.515 & -- \\
\midrule
Ours ($k$-NN)                              & \underline{17.97} & \underline{4.50} & \underline{0.546} & 0.595 & \underline{14.02} & \textbf{3.67} & \textbf{0.658} & 0.171 \\
\textbf{Ours (rigid clust.)}               & \textbf{14.89} & \textbf{3.98} & \textbf{0.588} & \textbf{0.628} & \textbf{11.93} & \underline{4.00} & \underline{0.638} & \textbf{0.498} \\
\bottomrule
\end{tabular}
\vspace{2pt}
\begin{minipage}{\linewidth}\scriptsize $^\dagger$ProxyPose predicts a pose per query rather than a dense per-pixel field; we evaluate its single-query variant under its own protocol with ground-truth intrinsics, so it is reported at the object/part level only and has no predicted mask for IoU. We omit its ADD, whose mean is dominated by a few clips where its tracker diverges (Appendix~\ref{sec:appendix_eval_se3}).\end{minipage}
\end{table}

%% file: tables/dataset.tex
\providecommand{\cmark}{\ensuremath{\checkmark}}
\providecommand{\xmark}{\ensuremath{\text{\sffamily X}}}
\providecommand{\pmark}{\ensuremath{\circ}}
\begin{table}[h!]
\centering
\caption{\textbf{Comparison of supervision signals across point-tracking and articulation datasets.} The point of this table is \emph{which} annotation channels and scene regimes each dataset exposes, not raw scale. ``Phys.\ multi-body'' = multiple independently moving bodies that interact under physics simulation (collisions, contact). ``Articul.'' = bodies whose motion is a kinematic tree of multiple rigidly-moving links. ``3D tracks'' = dense 3D point trajectories over $>2$ frames. ``Per-link \SE'' = world-frame $4{\times}4$ pose for each sub-object rigid link, every frame; \pmark{} = only an object-level (single-rigid-body) pose. ``Rigid-grouping'' = per-pixel partition of pixels by shared transform, beyond instance segmentation. ``Cam.\ variety'' = multiple distinct camera trajectory modes (mixing moving and stationary), in the spirit of Kubric's MOVi setups. Among the datasets compared here, Art-Kubric is the only one that provides all seven properties.}
\label{tab:dataset_compare}
\small
\setlength{\tabcolsep}{4pt}

\resizebox{\linewidth}{!}{%
\begin{tabular}{lcccccccc}
\toprule
Dataset & Source & \shortstack{Phys.\\multi-body} & \shortstack{Articul.\\(multi-link)} & \shortstack{Long-range\\tracks} & \shortstack{3D track\\GT} & \shortstack{Per-link\\\SE\,GT} & \shortstack{Rigid-\\grouping} & \shortstack{Cam.\\variety} \\
\midrule
FlyingThings3D~\cite{mayer2016large}             & Synth & \xmark & \xmark & \xmark & \cmark$^\flat$ & \xmark & \xmark & \cmark \\
Kubric MOVi-F~\cite{greff2022kubric}              & Synth & \cmark & \xmark & \cmark & \cmark$^\sharp$ & \pmark & \xmark & \cmark \\
TAP-Vid Kubric~\cite{doersch2022tap}              & Synth & \cmark & \xmark & \cmark & \xmark & \xmark & \xmark & \cmark \\
PointOdyssey~\cite{zheng2023pointodyssey}         & Synth & \pmark$^\diamondsuit$ & \xmark$^\dagger$ & \cmark & \cmark & \xmark & \xmark & \cmark \\
Dynamic Replica~\cite{karaev2023dynamicstereo}    & Synth & \xmark$^\dagger$ & \xmark$^\dagger$ & \cmark & \cmark & \xmark & \xmark & \cmark \\
SynthVerse~\cite{zhao2026synthverse}              & Synth & \xmark$^\ddagger$ & \cmark$^\ddagger$ & \cmark & \cmark & \xmark & \xmark & \xmark$^\ddagger$ \\
ARCTIC~\cite{fan2023arctic}                       & Real  & \xmark & \cmark & \pmark$^\star$ & \pmark$^\star$ & \cmark & \pmark$^\star$ & \xmark \\
HOI4D~\cite{liu2022hoi4d}                         & Real  & \xmark & \cmark & \pmark$^\star$ & \xmark & \cmark & \xmark & \xmark \\
\midrule
\textbf{Art-Kubric} (ours)                & Synth & \cmark & \cmark & \cmark & \cmark & \cmark & \cmark & \cmark \\
\bottomrule
\end{tabular}
}
\smallskip
{\footnotesize $^\flat$\,frame-pair scene flow only, not long-range. $^\sharp$\,not released as track files; computable from depth\,$+$\,per-instance pose. $^\dagger$\,foreground content is dominated by skinned (not piecewise-rigid) humans/animals; the ``Articul.'' column requires kinematic trees of \emph{rigidly-moving} links, which skinned characters do not satisfy. Dynamic Replica's backgrounds are static scanned environments and have no rigid-body physics. $^\diamondsuit$\,PointOdyssey applies random forces with realistic collisions on auxiliary GSO/PartNet rigid objects, but this multi-body physics is secondary to the character animation that dominates each scene. $^\ddagger$\,SynthVerse's articulated split renders one articulated asset per scene in an HDR surround (no other dynamic objects), drives motion by random joint-angle sampling within valid limits rather than by interaction forces (executed in Isaac Sim), and uses a fixed four-view rig plus one orbiting camera (\cite{zhao2026synthverse}, \S 3.3). The articulated source pool combines PartNet-Mobility~\cite{xiang2020sapien,mo2019partnet} and Infinite Mobility~\cite{lian2025infinite}; released annotations contain only point-track fields and instance masks (no per-link \SE). $^\star$\,not packaged as track/grouping files; derivable from released per-frame mesh\,$+$\,per-part 6D pose. ARCTIC's hands are MANO-parametric (not piecewise-rigid), so its rigid-grouping covers only the two rigid object parts.}
\end{table}